\documentclass[sigconf]{acmart}
\copyrightyear{2019} 
\acmYear{2019} 
\setcopyright{rightsretained} 
\acmConference[KDD '19]{The 25th ACM SIGKDD Conference on Knowledge Discovery and Data Mining}{August 4--8, 2019}{Anchorage, AK, USA}
\acmBooktitle{The 25th ACM SIGKDD Conference on Knowledge Discovery and Data Mining (KDD '19), August 4--8, 2019, Anchorage, AK, USA}
\acmDOI{10.1145/3292500.3330944}
\acmISBN{978-1-4503-6201-6/19/08}

\usepackage{balance}
\def\BibTeX{{\rm B\kern-.05em{\sc i\kern-.025em b}\kern-.08emT\kern-.1667em\lower.7ex\hbox{E}\kern-.125emX}}
    
\usepackage[utf8]{inputenc} % allow utf-8 input
\usepackage[T1]{fontenc}    % use 8-bit T1 fonts
\usepackage{hyperref}       % hyperlinks
\usepackage{url}            % simple URL typesetting
\usepackage{booktabs}       % professional-quality tables
\usepackage{amsfonts}       % blackboard math symbols
\usepackage{nicefrac}       % compact symbols for 1/2, etc.
\usepackage{microtype}      % microtypography
\usepackage{amsmath}
\usepackage{amsfonts}
\usepackage{amssymb}
\usepackage{subcaption,verbatim,engord,enumitem,multirow,bm,caption}
\usepackage{graphicx}
\usepackage{wrapfig}
\usepackage{algorithm,algorithmic}

\newtheorem{definition}{Definition}

\begin{document}

\title{Quantifying Long Range Dependence in Language\\
and User Behavior to improve RNNs}

\author{Francois Belletti}
\email{belletti@google.com}
\affiliation{%
  \institution{Google Research}
  \streetaddress{1600 Amphitheatre Parkway}
  \city{Mountain View}
  \state{CA}
  \postcode{94043}
}

\author{Minmin Chen}
\email{minminc@google.com}
\affiliation{%
  \institution{Google Research}
  \city{Mountain View}
  \state{CA}
  \postcode{94043}
}

\author{Ed H. Chi}
\email{edchi@google.com}
\affiliation{%
  \institution{Google Research}
  \city{Mountain View}
  \state{CA}
  \postcode{94043}
}

%
% By default, the full list of authors will be used in the page headers. Often, this list is too long, and will overlap
% other information printed in the page headers. This command allows the author to define a more concise list
% of authors' names for this purpose.
\renewcommand{\shortauthors}{Belletti et al.}

\begin{abstract}
    Characterizing temporal dependence patterns is a critical step in understanding the statistical properties of sequential data. 
    Long Range Dependence (LRD) --- referring to long-range correlations decaying as a power law rather than exponentially w.r.t. distance --- demands a different set of tools for modeling the underlying dynamics of the sequential data. 
    %Long Range Dependence (LRD) is a key feature of sequential datasets and helps categorize dynamical systems.
    While it has been widely conjectured that LRD is present in language modeling and sequential recommendation, the amount of LRD in the corresponding sequential datasets has not yet been quantified in a scalable and model-independent manner.
    We propose a principled estimation procedure of LRD in sequential datasets based on established LRD theory for real-valued time series and apply it to sequences of symbols with million-item-scale dictionaries.
    In our measurements, the procedure estimates reliably the LRD in the behavior of users as they write Wikipedia articles and  as they interact with YouTube.
    We further show that measuring LRD better informs modeling decisions in particular for RNNs whose ability to capture LRD is still an active area of research.
    % Capturing LRD patterns has recently been central to a large body of sequential models such as Recurrent Neural Networks (RNNs).
    The quantitative measure informs new Evolutive Recurrent Neural Networks (EvolutiveRNNs) designs, leading to state-of-the-art results on language understanding and sequential recommendation tasks at a fraction of the computational cost.
\end{abstract}

% TODO: change this!
\begin{CCSXML}
<ccs2012>
<concept>
<concept_id>10002950.10003648.10003688.10003693</concept_id>
<concept_desc>Mathematics of computing~Time series analysis</concept_desc>
<concept_significance>500</concept_significance>
</concept>
<concept>
<concept_id>10002951.10003317.10003331.10003271</concept_id>
<concept_desc>Information systems~Personalization</concept_desc>
<concept_significance>300</concept_significance>
</concept>
<concept>
<concept_id>10002951.10003317.10003338.10003341</concept_id>
<concept_desc>Information systems~Language models</concept_desc>
<concept_significance>300</concept_significance>
</concept>
<concept>
<concept_id>10010147.10010257.10010293.10010319</concept_id>
<concept_desc>Computing methodologies~Learning latent representations</concept_desc>
<concept_significance>300</concept_significance>
</concept>
</ccs2012>
\end{CCSXML}

\ccsdesc[500]{Mathematics of computing~Time series analysis}
\ccsdesc[300]{Information systems~Personalization}
\ccsdesc[300]{Information systems~Language models}
\ccsdesc[300]{Computing methodologies~Learning latent representations}

\keywords{data mining; sequential learning; long range dependence; RNN}

\maketitle

\section{Introduction}
Sequential modeling based on state-of-the-art Machine Learning (ML) techniques such as RNNs has proven very successful for applications such as natural language processing~\cite{mikolov2010recurrent,sutskever2014sequence,bahdanau2014neural,cho2014learning,bai2018empirical}, machine translation~\cite{cho2014learning}, speech recognition~\cite{graves2013speech}, and  recommender systems~\cite{hidasi2015session,smirnova2017contextual,wu2017recurrent,belletti2018factorized}.
As sequential modeling uses previously observed data to predict future states and observations, the rate at which past information becomes less relevant to present dynamics is a key aspect of the data. 
LRD focuses on sequences where past information losses its influence with a ``power-law'' decay --- as opposed to exponential --- which corresponds to many real-world time series (in finance, network traffic prediction, behavioral analysis~\cite{pipiras2017long}, human genomes, musics and languages).
% LRD is a critical characteristic in a dynamical system as higher LRD implies that perturbations have a longer lasting footprint which does not relax at a memory-less exponential rate.
The presence of LRD in sequential data dictates modeling choices, as higher LRD implies that perturbations have a longer lasting footprint which requires sequential models with longer memories.

While principled theories of LRD exist to analyze datasets, build models and prove limit theorems for real-valued time series~\cite{doukhan2002theory,samorodnitsky2007long,pipiras2017long,beran2017statistics}, the counterparts for sequences of discrete symbols are not readily available~\cite{lin2016criticality,belletti2018factorized,tang2019towards}, especially
when the set of symbol values (e.g. number of words in the English language) is large.
Being able to quantify the amount of LRD in such sequences can help better understand the statistical properties of the sequential dynamics and inform modeling decisions. 
  %  in user interactions with online platforms.
For instance, a more precise understanding of temporal dependencies in the data can guide the architectural design for language understanding and recommender systems --- two major areas of application of ML in production systems --- that has to conciliate two seemingly antithetic goals: leveraging long term history to better inform the next prediction and serving the prediction under a short latency deadline.

RNNs constitute the current state-of-the-art solution for sequential language and user behavior modeling~\cite{miller2018recurrent,chen2018dynamical,chang2017dilated,quadrana2017personalizing,devooght2017long,belletti2018factorized} and are widely employed in production systems.
RNNs are parametric non-linear models defined by a recurrent equation involving real-valued vectors:
\begin{equation}
    \label{eq:standard_rnn}
    [\hat{Y}_{t}, M_{t}]^T = \Phi_\theta\left(X_{t}, M_{t-1}\right)
\end{equation}
where $X_t$ and $\hat{Y}_t$ are the input and output at each timestep $t$ and $M_{t}$ is the state of the RNN storing past information. 
The ability of such networks to capture LRD patterns remains the subject of active research, with a large body of work focusing on improving the trainability of RNNs~\cite{pascanu2013difficulty,belletti2018factorized,miller2018recurrent,chen2018dynamical,trinh2018learning} in the LRD setting.

Enabling LRD in sequential neural models has been actively researched, in particular for language understanding~\cite{miller2018recurrent,chen2018dynamical,chang2017dilated} and sequential recommendations~\cite{quadrana2017personalizing,devooght2017long,belletti2018factorized,tang2019towards}, but remains challenging.
% In both these fields, accounting for the memory humans present in their sequential activities remains a major challenge.
Better accounting for long term user memory provides better behavioral predictions however LRD models are often difficult to serve in latency-sensitive production environments and at the scale of an entire online platform.
We show in the following that a better quantitative measurement of LRD leads to a finer understanding of behavioral dynamics which in turn enables more subtle trade-offs between model expressiveness and computational cost. 
The present paper improves the understanding of LRD in sequential behavioral modeling as follows:
\begin{enumerate}
    \item We provide an estimation method for LRD patterns in sequences of symbols belonging to large vocabularies, typical of language modeling and sequential recommendation tasks. To the best of our knowledge, we are the first to offer a model-free estimation of such dependencies;
    \item We show how the quantitative insights we extract can be turned into design decisions for RNNs operating under tight latency constraints. We propose a class of RNNs named Evolutive RNNs (EvoRNNs) which vary model capacity over time to match the sequential dependency patterns;
    \item We demonstrate the computational efficiency as well as quality gain of EvolutiveRNNs on the Billion Word language modeling task and an anonymized proprietary data set used to improve sequential recommendations for YouTube.
\end{enumerate}

\begin{figure}
    \centering
    \includegraphics[trim=0cm 0cm 1cm 0cm, width=0.5\linewidth]{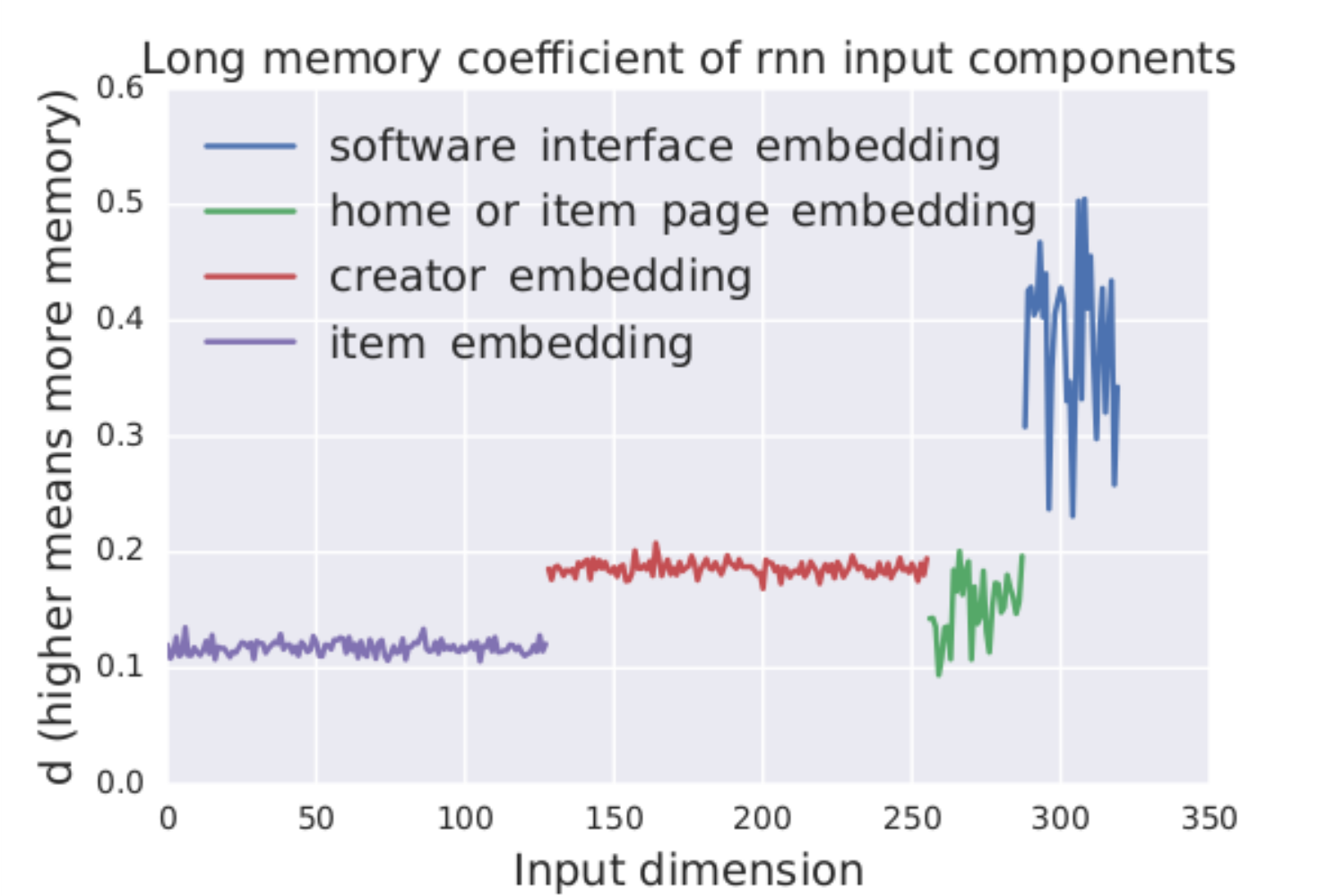}
    \includegraphics[width=0.45\linewidth]{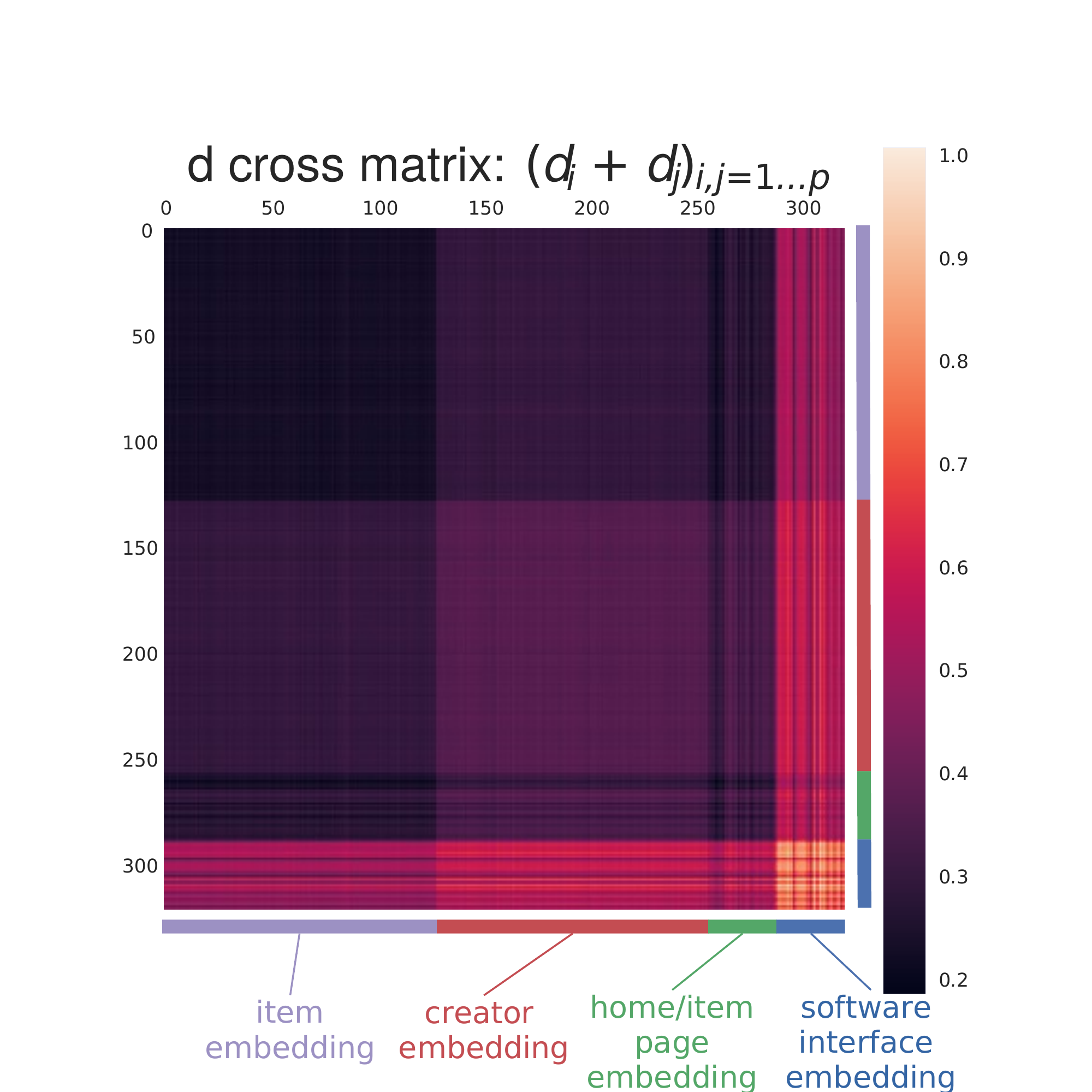}
    \caption{
    \footnotesize{
    Estimating Long Range Dependence (LRD) in user/item interactions on YouTube: 
    we measure the LRD coefficient (i.e. statistical sequential memory) in user behavior.
    The estimated LRD coefficients correspond to four groups of symbol embeddings.
    We present univariate memory coefficient (top) and covariate memory coefficients (bottom).
    For a large distance $h$, the component $i, j$ of the auto-covariance matrix $\gamma_X(h)$ decays proportionally to $h^{d_i + d_j - 1}$.
    A higher coefficient $d$ means that a higher amount of memory is present and $d$ significantly higher than $0$ demonstrates that the time series under consideration is Long Range Dependent.
    The estimates are obtained by applying the log-periodogram regression technique to sequences of symbols translated into their embeddings.
    }}
    \label{fig:d_estimates_seq_rec}
\end{figure}

\section{Related work}
\subsection{LRD in real-valued time series}
We follow the presentation of LRD given in~\cite{pipiras2017long} and consider a second-order stationary real-valued time series
$(X_t)_{t \in \mathbb{Z}}$, abbreviated to $(X)$.
By assumption, the mean of the time series $\mu_X = E[X_t]$ and its auto-covariance function 
$\gamma_X(h) = Cov(X_t, X_{t + h})$ 
are well defined and do not change over time.
The spectral density $f_X$ of $(X)$ is also well defined and verifies 
$\int_{-\pi}^{\pi} f_X(\lambda) e^{i h \lambda} d\lambda = \gamma_X(h)$. Prior to delving in the topic of LRD, we recall the definition of slow varying functions, \textit{e.g.}, the logarithm.

\begin{definition}{\textbf{Slow varying function:}}
A function $L$ is slow varying at infinity if it is positive on some interval $[c, \infty)$ where $c \geq 0$
and for any $a > 0$
$$
\lim_{u \rightarrow \infty} \frac{L(au)}{L(u)} = 1.
$$
A function $L$ is slow varying near $0$ if $u \rightarrow L(\frac{1}{u})$ is slow varying at infinity.
\end{definition}

Five different non-equivalent definitions of LRD are given in~\cite{pipiras2017long}. Here we only consider two equivalent definitions given respectively in the time and frequency domain.

\begin{definition}{\textbf{Long Range Dependent (LRD) mono-variate real-valued time series:}\label{def:mono-LRD}}
The mono-variate real-valued time series $(X)$ is LRD iff.
there exists a real $d \in (0, \frac{1}{2})$, referred to as the \textbf{LRD coefficient of $(X)$}, such that 
$$
    \gamma_X(h) = L_{\infty}(h) h^{2d - 1}
    \text{, or equivalently }
    f_X(\lambda) = L_{0^+}(\lambda) \lambda^{-2d},
$$
where $L_{\infty}$ and $L_{0^+}$ are slow varying functions at infinity and $0$ respectively.
\end{definition}

Higher value of $d$ indicates a slower decay of temporal dependence and therefore a higher amount of memory in the time series. Some readers may be acquainted with the Hurst index $H$ which measures the amount of LRD in a stochastic process through its self-similarity and scaling properties~\cite{pipiras2017long,mandelbrot1998fractals,sornette2006critical}. If $H \in (\frac{1}{2}, 1)$ then $H$ verifies $d = H - \frac{1}{2}$ (this property is for instance proven for Fractional Brownian Motions in~\cite{pipiras2017long}).

A consequence of the time series $(X)$ being LRD is that the variance of 
$\frac{\Sigma_{t=1}^N X_t}{N}$ decays much slower as $L_{\infty}(N) N^{2d - 1}$ where $L_{\infty}$ is another slow varying function at infinity.
LRD is indeed notorious for changing convergence rates of $M$ estimators as compared to the case of \emph{iid}. observations~\cite{doukhan2002theory,samorodnitsky2007long,pipiras2017long,beran2017statistics}.

As explained in~\cite{doukhan2002theory,samorodnitsky2007long,pipiras2017long,beran2017statistics} there are multiple standard estimators for the LRD coefficient $d$ such as the Rescaled Range estimator $R/S$, wavelet based, and variance estimation based estimator. Maximum Likelihood Estimators for generative linear LRD models such as FARIMA models are also available.
One long-standing estimator for $d$ is the log-periodogram estimator~\cite{robinson1995log} which focuses on the spectral density
$
    f_X(\lambda) = L_{0^+}(\lambda) \lambda^{-2d}
$.
Let $\widehat{f_X(\lambda)} \equiv |FFT_N[\lambda](X)|^2 = |\sum_{t=1}^N X_t e^{-it\lambda}|^2$ denote the empirical spectrum of $(X)$ --- assuming $N$ observations of the time series are available --- then one can measure $d$
through the estimation of the slope $b=-2d$ in the affine relationship
\begin{equation}\label{eq:log-periodogram}
    %\log \left( |FFT_N[\lambda](X)|^2 \right) = a + b \log(\lambda)
    \log \left( \widehat{f_X(\lambda)}\right) = a + b \log(\lambda)
\end{equation}
by ordinary least squares regression in the domain of low frequencies~\cite{robinson1995log}.

Although Maximum Likelihood Estimation is now preferred for measuring LRD in time series~\cite{pipiras2017long}, 
we employ the log-periodogram estimate here to avoid assuming a particular generative model for the data.
Therefore, we propose to methodically quantify LRD in sequences of symbols in large vocabularies/inventories through the spectral density of sequences.

\subsection{LRD in sequences of symbols}
LRD often manifests itself in physical and societal phenomena through a slow decay of temporal dependence
which is usually observed in the form of a power-law decaying auto-covariance function~\cite{pipiras2017long}. Per Definition~\ref{def:mono-LRD}, this time domain power-law decay at infinity is equivalent to a power-law divergence of the spectral density in the frequency domain near $0$.

LRD estimation has become standard in the study of real-valued time series and has led to improvements in LRD predictions or risk assessment thanks to models such as FARIMA~\cite{pipiras2017long,sornette2006critical}. 
In contrast, while it is widely assumed that LRD is a key feature of  the input sequences that needs to be captured for better predictions in language modeling and sequential recommendations, the amount of LRD in these tasks remains to be estimated in a principled manner. 

A key difference between real-valued time series and language modeling or sequential recommendation tasks
is that the latter generally involve sequences of discrete symbols or items from vocabularies of $10^5$ to $10^7$ distinct values.
For small vocabularies of symbols, computing the decay of mutual information along the time axis helps quantify the amount of LRD as in sequences of characters~\cite{lin2016criticality}. Unfortunately, these techniques do not scale to large vocabularies due to sparse observations and the prohibitively large number of possible combinations. In the present paper we show how alternate representations of symbols can scale estimates of the LRD coefficient $d$ to sequences involving large vocabularies of symbols.

\subsection{Gradient propagation and LRD in RNNs}
Model-free estimators of LRD such as the log-periodogram estimator differ radically from usual measures of LRD employed in sequential neural models.  A substantial body of work concerned with the application of RNNs to LRD sequences of inputs focuses on the propagation of gradients through time.

From the seminal paper on the difficulty of training RNNs~\cite{pascanu2013difficulty} to recent developments~\cite{belletti2018factorized,miller2018recurrent}, exploding or vanishing gradients are considered the main obstacle to LRD modeling in RNNs. Various approaches have been proposed to address the issues.  Modifications started by introducing gating as in LSTMs~\cite{hochreiter1997long} and GRUs~\cite{chung2014empirical} and later by building multi-scale temporal structure~\cite{chung2016hierarchical,chang2017dilated}, constraining on the spectrum of learned parameter matrices~\cite{arjovsky2016unitary,jing2016tunable,vorontsov2017orthogonality}, regularization~\cite{trinh2018learning,merity2017regularizing} and initialization schemes~\cite{chen2018dynamical} to improve the trainability of RNNs. It is worth mentioning here that RNNs are not the only neural models to provide good performance with long sequences of inputs. For instance, dilated convolutional architectures~\cite{van2016wavenet,yu2015multi} have been offered as an effective alternative. 
Attention is also readily able to capture LRD patterns as part of Transformer~\cite{vaswani2017attention} but unfortunately it is challenging to serve such a multi-layer attention network with the very low latency required by recommender systems.
An alternate solution may be to use a single attention layer as part of a Mixture-of-Experts~\cite{tang2019towards}.
%are also provably effective.

% Such approaches in fact measure the ability of a given model to capture LRD instead of the amount of LRD in input sequences itself.
% It is worth signaling that Information Theoretical analysis has also been successful in helping understand the trainability and capacity of RNNs~\cite{collins2017capacity} but this analysis focused on the model and not the data itself.

% Considering LRD to be a key feature of sequential patterns in language modeling and sequential recommendations or in synthetic tasks requiring long memory~\cite{pascanu2013difficulty} has led to substantial transformations of RNNs.
% Modifications started by creating separate memory units in LSTMs~\cite{hochreiter1997long} and GRUs~\cite{chung2014empirical} and later involved multi-scale temporal analysis~\cite{yu2018sliced,chung2016hierarchical,dieng2016topicrnn,mujika2017fast,chang2017dilated}, imposed constraints on the spectrum of learned parameter matrices~\cite{arjovsky2016unitary,jing2016tunable,dangovski2017rotational,vorontsov2017orthogonality}, focused on regularization~\cite{trinh2018learning,merity2017regularizing} or initialization schemes~\cite{chen2018dynamical} to improve the trainability of RNNs. It is worth mentioning here that RNNs are not the only neural models to provide good performance with long sequences of inputs. For instance, dilated convolutional architectures such as~\cite{van2016wavenet,yu2015multi} which can be interpreted as generalizations of wavelet transforms~\cite{mallat2008wavelet} are also provably effective.

The key novelty of our approach is to not measure LRD as the propagation of information through a RNN but rather estimate LRD in the sequences of inputs themselves and design model architectures to match the dependence patterns. 
Therefore, although we mostly focus on architectural insights for RNNs, our approach is in no way limited to this class of models and could inform the design of other sequential models such as convolution or attention~\cite{bahdanau2014neural} based neural architectures.

\section{Estimation of LRD with distributed word and item representations}
\label{sec:estimate}
The lack of model-independent measures of LRD blurs the distinction between the amount of LRD present in input sequences and the ability of neural sequential models to capture such LRD.
To fill this lacuna, we now develop an estimate for the LRD coefficient $d$ as defined in Definition~\ref{def:mono-LRD} for long sequences of symbols with large vocabularies, which are common in language modeling and sequential recommendation tasks.

\subsection{From sequences of discrete symbols to real valued embedding time series}
%\subsection{Embedding symbols from large dictionaries}
For tasks involving sequences of symbols belonging to large vocabularies ($10^4$ to $10^7$ in size), it has become standard practice to embed each item in the form of learned vector of real values
%~\cite{sutskever2014sequence,bahdanau2014neural,mikolov2010recurrent,cho2014learning,bai2018empirical,tang2018personalized,smirnova2017contextual,covington2016deep,wu2017recurrent,quadrana2017personalizing,devooght2017long,belletti2018factorized,yu2015multi,miller2018recurrent,dieng2016topicrnn,chung2016hierarchical,mujika2017fast,chang2017dilated,chen2018dynamical,hidasi2015session,sarwar2001item,linden2003amazon} 
since the seminal work on word2vec~\cite{mikolov2013distributed,mikolov2013linguistic} and Glove embedding~\cite{pennington2014glove}.

These methods map discrete symbols to real-valued vectors in $\mathbb{R}^p$. In practice, a few hundred embedding dimensions are sufficient to provide state-of-the-art predictive performance for tasks with vocabularies of several millions of symbols~\cite{mikolov2013distributed,covington2016deep,belletti2018factorized}.  Close examination of the inter-item relationships inherited from these continuous representations~\cite{mikolov2013distributed,mikolov2013linguistic,maaten2008visualizing,xin2017folding} suggests that related items are indeed collocated in the embedding space.  

With these embeddings, we can map sequences of discrete symbols to sequences of real-valued vectors, and use methods developed for real-valued multi-variate time series for analysis. In particular, the well established theory of LRD~\cite{pipiras2017long} can be used to characterize the sequential dependence properties of sequences of learned item embeddings.
% Symbol embeddings constitute a collection of learned parameters in $\mathbb{R}^p$ and are typing optimized as part of the training of the whole model for the task of interest.
% While certain dictionaries of symbols may have several million of items~\cite{covington2016deep,belletti2018factorized} a few hundred embedding dimensions are still sufficient to provide state-of-the-art predictive performance. A body of literature focuses on interpreting~\cite{mikolov2013distributed,mikolov2013linguistic}, assessing~\cite{xin2017folding}, and visualizing~\cite{maaten2008visualizing} the inter-item relationships learned by such continuous representations for discrete symbols.
% It is indeed expected that related items should be collocated in the embedding space. As embeddings map discrete items to real-valued vector they pave the way towards examining sequences of symbols belonging to large dictionaries with methods that work provably well for real-valued multi-variate time series.
%\subsection{From sequences of discrete symbols to real valued time series}
While most existing work focuses on interpreting~\cite{mikolov2013distributed,mikolov2013linguistic}, assessing~\cite{xin2017folding}, and visualizing~\cite{maaten2008visualizing} inter-item relationships, to the best of our knowledge, we are the first to examine such relationships \emph{longitudinally along the time axis}.
% Symbol embeddings are generally analyzed statically~\cite{mikolov2013distributed} and inter-item relations are examined visually with dimensionality reduction techniques such as t-SNE~\cite{maaten2008visualizing}.
% To the best of our knowledge, no previous work has examined such relationships \emph{longitudinally along the time axis}. 
% The key novel insight of the present work is to analyze the temporal dynamics of sequences embedding vectors as they turn series of discrete symbols into real-valued multi-dimensional time series. Many principled statistical methods are available to analyze sequences of real-valued vectors. In particular, the well established theory of LRD~\cite{doukhan2002theory,samorodnitsky2007long,pipiras2017long,beran2017statistics} can be used to characterize the sequential dependence properties of sequences of learned item embeddings.

\subsection{Estimation methods for LRD}
Although we have exposed the definition of the LRD coefficient $d$ in the mono-variate setting, we still need to extend the presentation to multi-variate time series as item embeddings are real-valued vectors. Here  we again follow the presentation given in~\cite{pipiras2017long}.

Consider a multi-variate second-order stationary time-series $(X_t)$ with $X_t \in \mathbb{R}^p$.
We denote $\gamma_X(h) = Cov(X_t, X_{t+h})$ the matrix-valued auto-covariance function of $(X)$ which takes values in $\mathbb{R}^{p\times p}$ and
$f_X$ the corresponding spectral density matrix which also takes values in $\mathbb{R}^{p\times p}$.
By definition $\gamma_X$ and $f_X$ satisfy
$
\forall j, k\in {1 \dots p}, \; \forall h \in \mathbb{Z}, \;$
$$\gamma_{X, j, k}(h) = \int_{-\pi}^{\pi} f_{X, j, k}(\lambda) e^{i h \lambda} d \lambda.
$$

\begin{definition}{\textbf{Long Range Dependent (LRD) multi-variate real-valued time series:}\label{def:multi_LRD}}
The multi-variate real-valued time series $(X_t)_{t \in \mathbb{Z}}$ is LRD iff.
there exists a real vector $(d_i)_{i = 1 \ldots p} \in (0, \frac{1}{2})^p$
such that
$$
    \gamma_{X,j, k}(h)(\lambda) = L^{j, k}_{+\infty}(h)\;h^{d_j + d_k - 1}
$$
or equivalently
$$
    f_{X, j, k}(\lambda) = L^{j, k}_{0+}(\lambda) \;\lambda^{-\left(d_j + d_k\right)}
$$
where 
$L^{j, k}_{+\infty}$ are slow varying functions at infinity and
$L^{j, k}_{0+}$ are slow varying functions close to $0+$.
\end{definition}

As a result, each element $j, k$ of the spectral density matrix of a multi-variate LRD time series can be written as
$
    f_{X, j, k} \sim g_{j, k} \lambda^{-(d_j + d_k)}
$
for low frequencies $\lambda$.
Similar to the mono-variate case, we can use a log-periodogram regression in low frequencies as a way to estimate $d$.

\subsection{LRD and Mutual Information}
Assuming $(X)$ is Gaussian, a relation between the rate of decay of the auto-correlation of $(X)$ and that of the Mutual Information $I\left(X_t, X_{t+h}\right)$ can be established~\cite{cover2012elements,mackay2003information}. One can easily prove that the mutual information of two multi-variate Gaussian random variables $U, V$ is
$$I(U; V) = \frac{1}{2}\log\left(\frac{\det(\sigma_U) \det(\sigma_V)}{\det(\sigma)}\right)$$
where $\sigma_{U}$ and $\sigma_{V}$ are the corresponding covariance matrices and $\sigma$ is the $2p\times 2p$ covariance matrix. 
Given that $(X)$ is second-order stationary and Gaussian, one can show that~\cite{guo2005additive,guo2005mutual}
\begin{align*}
    I & \left(X_t, X_{t+h}\right) - \log \det \left( 
            \gamma_X(0)
        \right)\\
    & = - \frac{1}{2} \left(
        \log \det \left (
            \gamma_X(0) \gamma_X(0) - \gamma_X(h) \gamma_X(h)
        \right)
    \right) \\
    & \sim \sum_{i=1}^p L^i_{+\infty}(h) h^{2 (2 d_i -1)}
\end{align*}
in the simple case where $\gamma_X$ is  diagonal. 
Here $(L^i_{+\infty})_{i=1 \dots p}$ are slow varying functions at infinity.
Therefore one can assume a characteristic power-law decay of the mutual information based on the values of $d$, which relates our spectral method for characterizing sequential memory to the mutual information based approach in~\cite{lin2016criticality}. In particular, in the mono-variate case, that is $p=1$, we have
$$\log I\left(X_t, X_{t+h}\right) \propto 2(2d-1) \log h.$$
That is, the slope of decay of mutual information w.r.t. separation in the log-log space corresponds to the LRD coefficient $d$.
Unfortunately in the multivariate case, the slope does not give access, even with the Gaussian diagonal assumption, to individual estimates of the components of $d$.

\subsection{Implementation at scale}
Let us first focus on a detailed algorithmic presentation of the estimation procedure we designed to estimate LRD coefficients in long sequences of symbols.

\begin{figure}
    \centering
    \includegraphics[width=\linewidth]{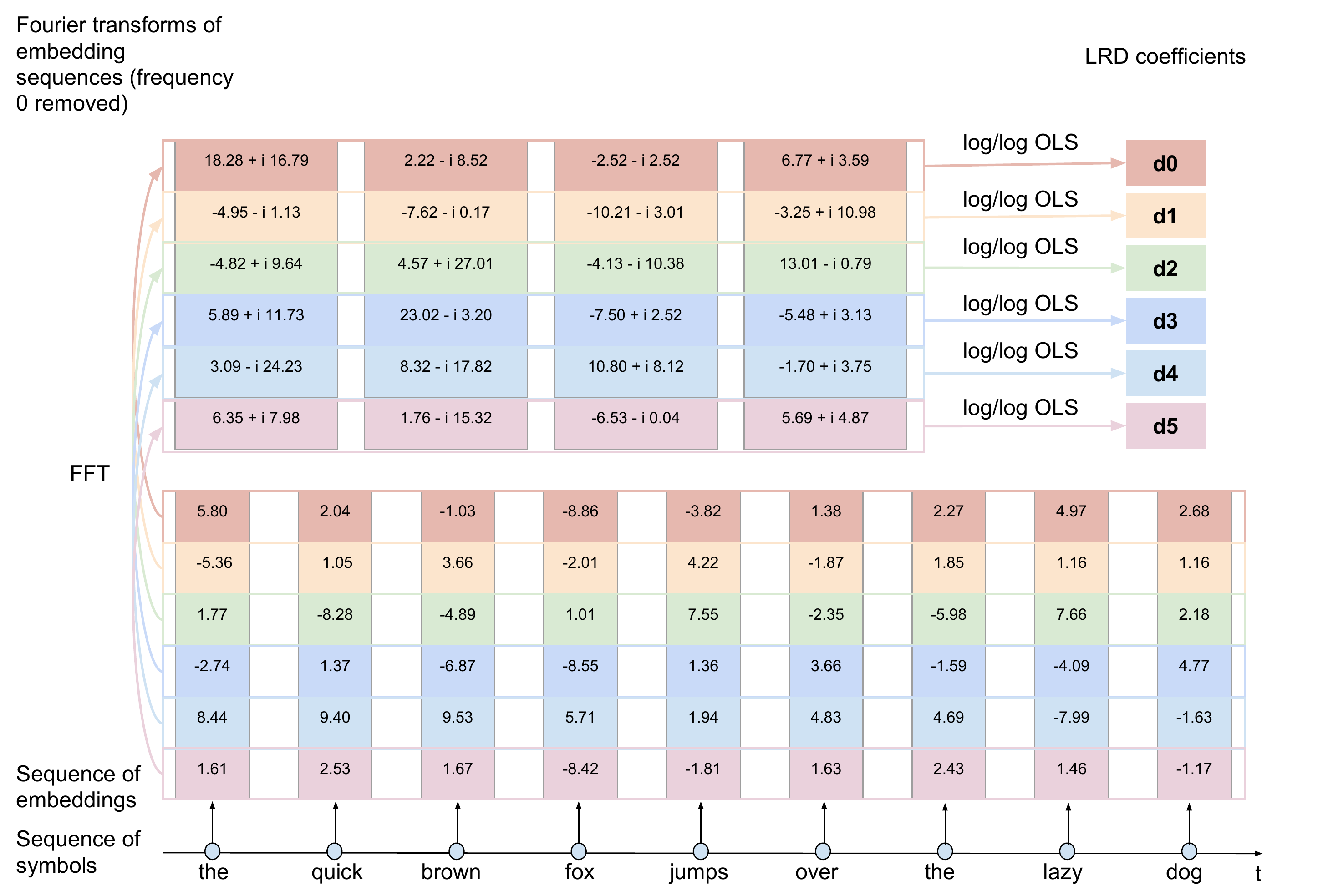}
    \caption{Schema of the log-periodogram estimation procedure employed in our study}
    \label{fig:schema_procedure}
\end{figure}

Algorithm~\ref{alg:OLS_procedure} details the actual implementation for a given sequence  and Figure~\ref{fig:schema_procedure} presents it schematically.
In order to scale the method to data sets comprising millions of sequences we run the procedure in a mini-batched manner, computing FFTs and OLSs in parallel, while the estimates for the coefficients $d$ update a global estimate with a chosen learning rate. More details on the implementation are given in appendix.

\begin{algorithm}
\caption{Estimate LRD coefficients for a sequence of symbols}
\label{alg:OLS_procedure}
\begin{algorithmic}
\REQUIRE{$L$ \COMMENT{padding length}, $p$ \COMMENT{embedding dimension}, $\mathbf{E}$ \COMMENT{symbol embeddings}}
\ENSURE{$d \in \mathbb{R}^p$}
\STATE{embeddingSequence $\gets$ lookupEmbedding($\mathbf{E}$, symbolSequence)}
\STATE{paddedEmbeddingSequence $\gets$ pad(embeddingSequence, $L$, $\mathbf{0}$) \COMMENT{pad the beginning of sequence with zero valued vectors to obtain a sequence of length $L$}}
\STATE{spectrum $\gets$ $\left| \text{RFFT}(\text{paddedEmbeddingSequence})[1:] \right|^2$ \COMMENT{remove the frequency $0$ term}}
\FOR{$i \gets 0$ to $p-1$}                    
    \STATE{d[i] $\gets$ OLS($\log$(range(1, $L$ // 2 + 1), $\log$(spectrum[:, i]))}
\ENDFOR
\STATE{return $d$}
\end{algorithmic}
\end{algorithm}

\subsection{Observations of LRD on actual sequential data sets}
We now apply the estimation of the memory coefficient vector $d$ to sequences of learned item embeddings in a language and a user-behavior dataset. 
It is worth pointing out that the log-periodogram estimate of LRD assumes that the time series are second-order stationary and the spectrum measurement using FFT uncovers only linear sequential dependency patterns. 
% There is no guarantee such time series are second-order stationary.
% Also, there is no reason a priori why the sequential dependency pattern should be linear.
% Although both these assumptions are key to our theoretical developments,
Although neither assumptions are guaranteed in any arbitrary time series, our method is sufficient to detect linear second order stationary LRD patterns if they exist, without guarantee that it will unravel any kind of non-linear or non-stationary LRD.
Our empirical results show that such a linear LRD does exist in the sequences of word embeddings and item embeddings corresponding respectively to text documents and user/item interactions on YouTube.
% We first consider long sequences of words for language understanding, and then user behavior data -- watched items on a content hosting platform for recommendation. % relying heavily of content recommendations.

\begin{figure}
    \centering
    \includegraphics[width=0.49\linewidth]{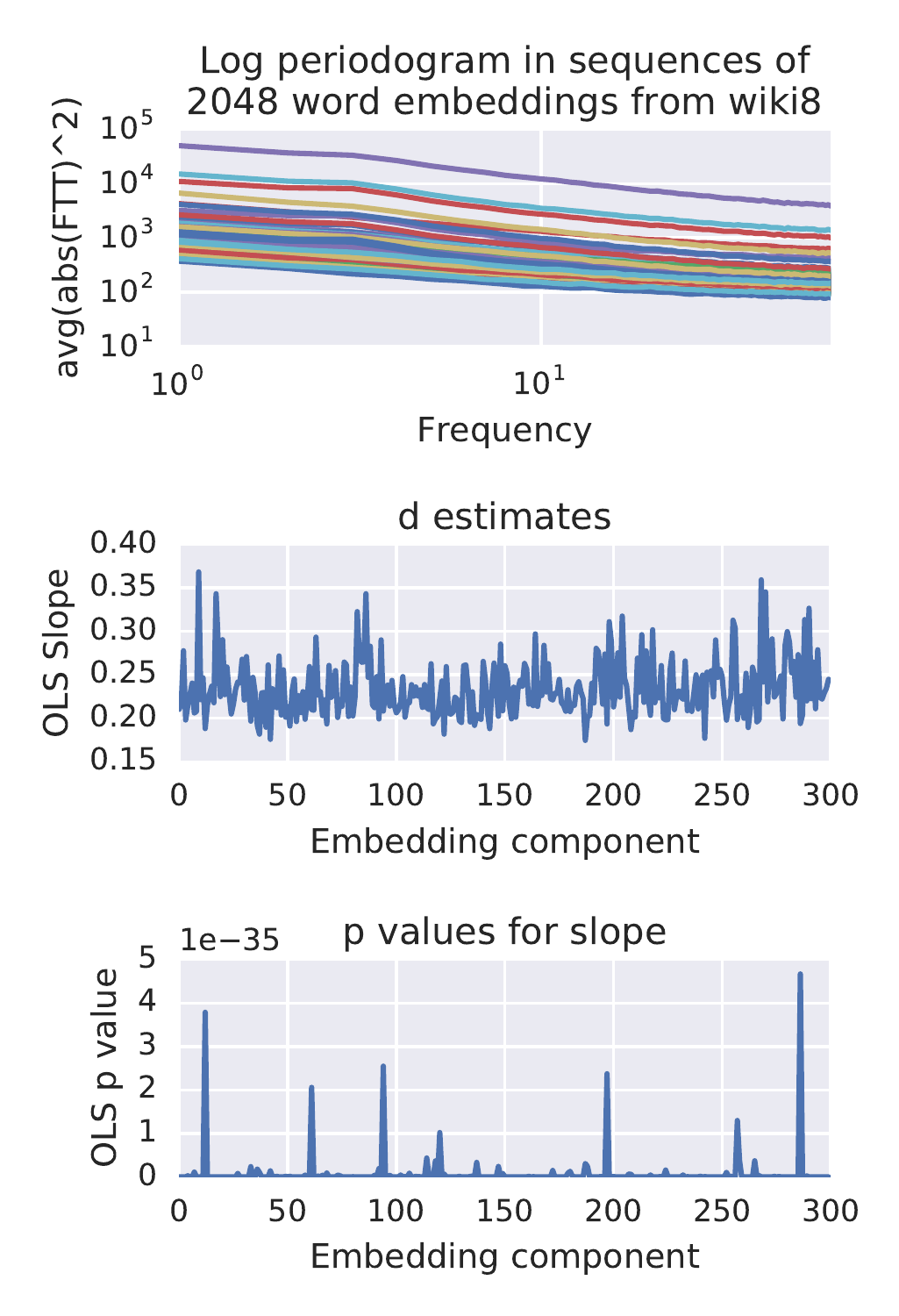}
    \includegraphics[width=0.49\linewidth]{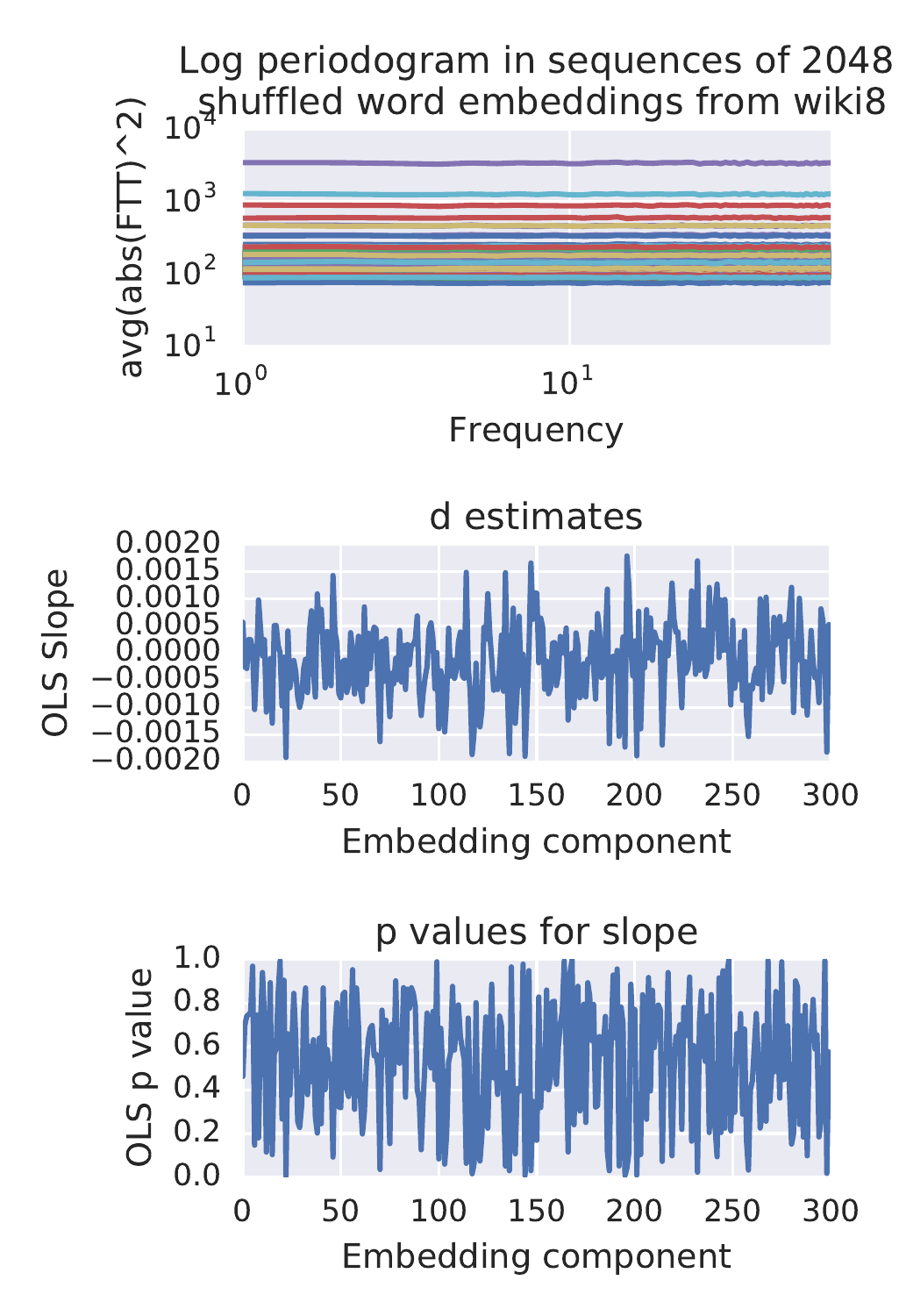}
    \caption{
\footnotesize{
Spectral density estimate (in log space), the LRD coefficients and the p-value for the OLS regression on the Wikipedia dataset (left). Similar estimation (right) with sequences whose words have been randomly shuffled.
    % Distribution of the LRD coefficients $(d_i)_{i=1 \dots p}$ as word embeddings are being learned through training. As expected, the noisy initial embeddings results in a low memory being estimated. Progressively, as the vector-valued item representations are being learned, certain component of the corresponding multi-variate time series become LRD. Here we use the pre-trained Glove embeddings and longer sequences (2048 items) to estimate spectra in the low frequency domain.
}
    \label{fig:LRD_lm1b}}
\end{figure}
\subsubsection{LRD in word sequences}\label{sec:estimate_lm}
We start with measuring the LRD coefficients $d$ on a subset of the Wikipedia dump consisting of concatenated Wikipedia articles (100 MB from Wikipedia) which keeps the sequential structure of the documents intact --- no processing is done besides removing punctuation, converting all letters to lowercase and removing other artifacts.
We break the documents into long sequences of 2048 words. 
Each sequence is then transformed into a multi-variate real-valued time series by mapping each word of the sequence to a pre-trained 300-dimensional Glove embedding~\cite{pennington2014glove}, learned on the 2014 Wikipedia dump. 
Next, we compute the average of the squared magnitude $\widehat{f_X(\lambda)}$ of the Fourier Transform of the sequence of word embeddings.
The slope of $\log\left(\widehat{f_X(\lambda)}\right)$ versus $\log(\lambda)$ on low frequencies is then estimated through OLS by minimizing the corresponding log-periodogram loss as shown in Equation~\ref{eq:log-periodogram}.
We tried different padding strategies for words whose embedding was unknown --- skipping, zero padding and mean learned embedding padding --- and did not find any substantial change in the resulting estimates of $d$.

We present the spectral density estimate, the LRD coefficients and the p-value for the OLS regression in Figure~\ref{fig:LRD_lm1b} (left). 
As shown in the figure, the spectral density estimates --- each curve corresponding to FFT of one dimension of the word embedding --- decay linearly near $0$ in the log-log space, and the estimates for the coefficients $d$, with all the 300 dimensions shown in the x-axis, are significantly positive.  These observations suggest that the time series under consideration is long range dependent according to the canonical statistical definition of LRD, which in return indicates that the input sequence is LRD. A higher coefficient demonstrates that a higher amount of memory is present. 
To give a confidence of our estimate, we also include the p-value of OLS estimating a slope of zero, that is $d = 0$. Extremely small p-values are returned, which again corroborates that the sequences are indeed LRD.
Ideally, one would tailor the p-value tests to the particular setting of OLS in log-scale (which violates some assumptions on the distribution of errors), but this is outside the score of the paper.
Detailed theory on the log-periodogram estimator can be found in~\cite{robinson1995log}.

We now proceed with a sanity check for the estimator of LRD through the log-periodogram method operating on vector-valued sequences of embedded symbols.
On Figure~\ref{fig:LRD_lm1b} (right) we include the estimates on the same data-sets with the words randomly shuffled within sequences of $2048$ words. Random shuffling dissolves any dependence patterns existing in the original data-set and gives a white-noise-like statistical structure to the embedding sequences. We can see that the spectral density no longer concentrate its mass around $0$ (low frequencies) and our log-periodogram estimator gives close to $0$ estimates for the LRD coefficient of the shuffled sequences with correspondingly high p-values. The side-by-side comparison showcased that our method is able to detect the presence of LRD (as in the original word sequences) vs not (as in the shuffled sequences). 
% In this shuffled case the method we propose does not conclude that LRD is present which was expected.

% The Billion word data set~\cite{chelba2013one} is a standard benchmark for language modeling~\cite{mikolov2010recurrent,bai2018empirical,chen2018dynamical} aimed at predicting the next word in a text.
% An architecture similar to the one shown in Figure~\ref{fig:estimation procedure} is employed for this predictive tasks and a vector embedding is learned for each word present in the dictionary. 
% We modified the baseline implementation of~\cite{jozefowicz2016exploring} and the sequence length is now $128$. Again we employ the log-periodogram estimator and present the resulting estimated values of the LRD coefficient $d$ in Figure~\ref{fig:LRD_lm1b}.
% We observe that the amount of LRD is noticeably lower in this second dataset.

% One key advantage of being able to quantify the amount of LRD is that it is now possible to compare the amount of memory needed to make accurate predictions on different datasets. We now use such data exploratory insights to inform neural model design.

\subsubsection{LRD in user behavior sequences}\label{sec:estimate_seq_rec}
% \subsubsection{Sequential recommendation data set}
Next we measure the LRD coefficients $d$ on user behavior data. 
The data set consists of user generated interaction sequences in a large-scale anonymized dataset from YouTube to which we have access through employment at YouTube working on improving YouTube for users.
% The dataset is private, anonymized and accessible only internally by few employees whose work is directly related to UGCPlatform.com.
Each sequence records a series of timestamped item ids corresponding to a given user $u$ starting to access an item $v$ at time $t_i$: $\left\{\xi^u_{t_i} \right\}_{i = 1 \dots N_u}$ --- where $N_u$ is the number of interactions available for user $u$. We clip the sequences to at most $500$ observations per user. This production data-set comprises of more than $200$ million training sequences, more than $1$ million test sequences and has an average sequence length of $200$.

%The data is used by a RNN based recommender to map a sequence of historical observations to the next item the user is likely to watch. The most likely items are then shown to the user in the form of impressions.

%Here we consider clipped sequences of at most $512$ observations per user. Shorter sequences are padded with zero once represented by embedded items. Note that such a zero padding is neutral in its contribution to the Fourier Transform of the embedded inputs.

% \begin{table}[t!]
% \center
% \caption{Sequential recommendation dataset.}\label{tb:seq_rec_dataset}
% \begin{tabular}{lc}
% \toprule
% \textbf{Production Dataset}\\
% \midrule
% Training sequences & >200M\\
% \midrule
% Test sequences & >1M\\
% \midrule
% Distinct items & $\approx$2M\\
% \midrule
% Avg. sequence length & $\sim200$\\
% \bottomrule
% \end{tabular}
% \end{table}

\begin{figure}
    \centering
    \includegraphics[width=0.8\linewidth]{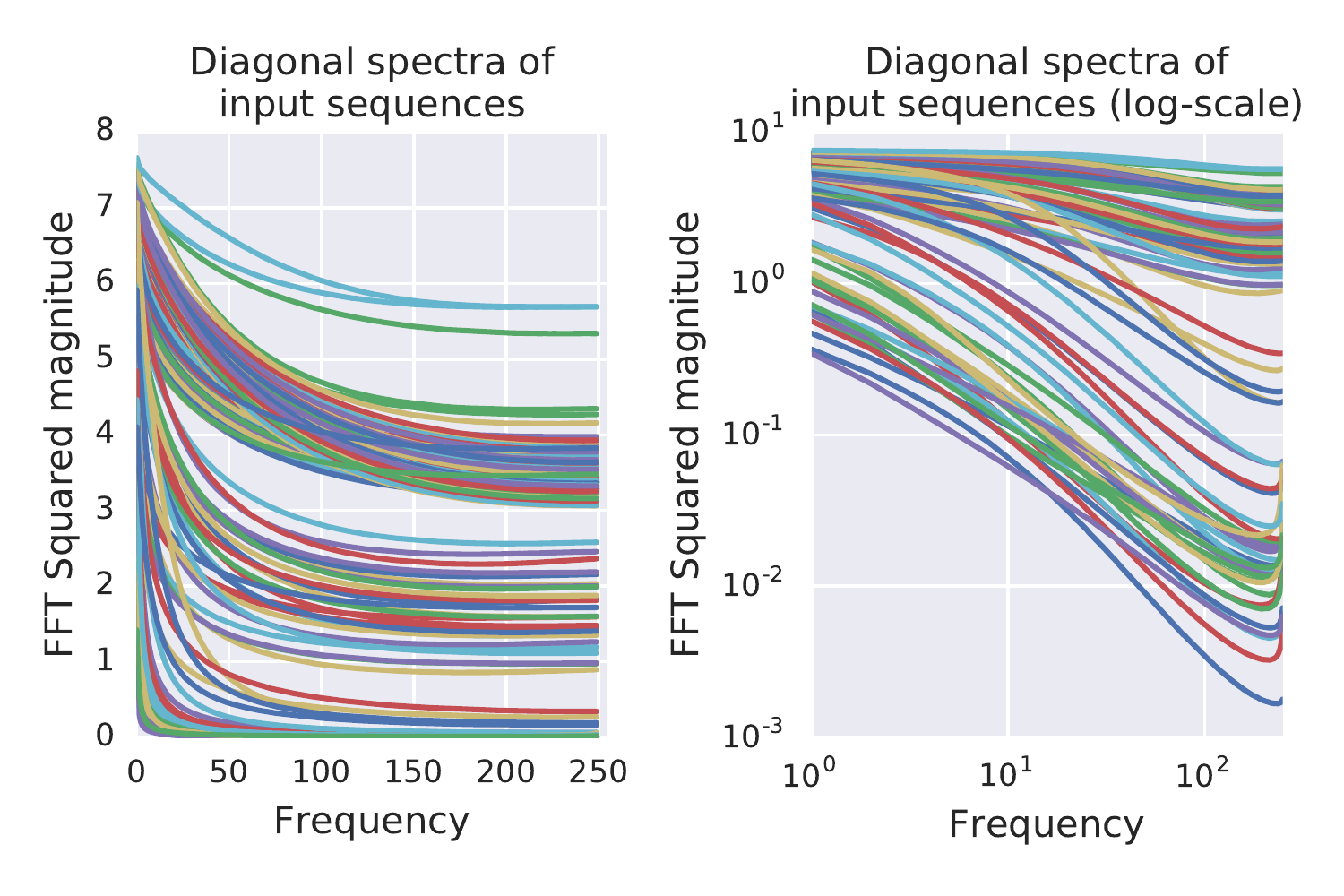}
    \caption{
\footnotesize{
    Spectral density of time series of embedded symbols in the original (left) and log-log space (right) where a linear decay is clear for several estimated spectra near $0$.
}}
    \label{fig:spectrum_seq_rec}
\end{figure}

%\subsubsection{Symbol embeddings}
Different from the word sequences where each $X_t$ involves a single symbol, $\xi^u_{t_i}$ includes multiple sub-symbols, each corresponding to one different aspect of the interaction: the item watched (from a vocabulary of 2 million items), the creator/publisher of the watched item (from a vocabulary of 1 million), the page the item was displayed (order of tens) and the OS employed by the user (orders of hundreds). The discrete values of these four groups of symbols are embedded with $128$, $128$, $32$ and $32$ dimensional real-valued vectors  %$\widetilde{v}$, $\widetilde{c}$, $\widetilde{p}$, $\widetilde{s}$ 
respectively. 
The embeddings are concatenated and trained as part of the sequential neural model aiming at predicting the next item the user will consume, which we are going to detail in section~\ref{sec:experiment_recommendation}. With the learned embeddings, we follow the same procedure as described in the word sequence case to estimate the LRD coefficients. Figure~\ref{fig:spectrum_seq_rec} plots the spectral density of the embedding sequences.  Again, the power law decay (left) and the linear decay in the log-log space (right) near $0$ are the clear marks of a LRD pattern. 
% An OLS linear regression in log-log scale estimates the power-law coefficients, as shown in Figure~\ref{fig:d_estimates_seq_rec}. 
Figure~\ref{fig:d_estimates_seq_rec} (left) shows the estimated four groups of coefficients $d$. 
% As the sequence of input embeddings is multi-variate with groups of dimensions in
% $([\widetilde{v}_t, \widetilde{c}_t, \widetilde{p}_t, \widetilde{s}_t])$
% representing different groups of symbols, we can examine the values of the different components of $(d_i)_{i=1 \dots p}$. Each component corresponds to a different aspect of the sequential behavior of users on the platform: the item watched, the corresponding creators, the web page, and the OS employed.
We notice that the embedding representing creators/publishers estimates larger coefficients $d$ than the embedding representing individual items, indicating more LRD.
Also, the software interface embedding carries the highest amount of LRD which is expected as it is less likely to change within short sequences of interactions. The maximum OLS p-value for the linear slopes in log-log scale being zero is $1.32 \times 10^{-108}$.

Another aspect in which these user-behavior sequences differ from the word sequences is the irregularly-spaced events. In this first work we do not take into account of that in order to use the Fast Fourier Transform algorithm readily when computing the spectral density of 
$(\widetilde{\xi}_t)$. % we consider the ordering of interactions and do not take into account the fact that the original timestamps $t_i$ are not regularly spaced. We employ such an event-based representation of temporal ordering to be able to use the Fast Fourier Transform algorithm readily when computing the spectral density of 
% $(\widetilde{\xi}_t)$.
The Fourier transform however is well defined for irregularly spaced time stamps~\cite{brillinger1981time,rahimi2008random,belletti2017random} and in future work we plan to apply dedicated estimators to improve our estimation under these settings.
% We also consider Language Modeling (LM) which naturally has regularly spaced observations and apply the same estimator on LM datasets as well.

\section{LRD and neural architecture design}
The estimation results above clearly indicate that there are long range dependence in sequences of words as well as sequences of user behavior data. 
%sequences of user generated interactions are LRD and while some components of word embeddings are LRD in the Billion Word datase, most are not. 
The presence of LRD implies a power law decay of auto-covariance between present  and past observation.
In other words, as opposed to short memory sequences~\cite{pipiras2017long} where the auto-covariance decays at an exponential rate, past information has a long lasting footprint.
Having confirmed the slow rate of decay of  relevance of past observations, we now change the architectural design of standard RNNs accordingly.

\textbf{Computational trade-offs and LRD.}
% Demonstrating the presence of LRD in input sequences poses a major challenge for sequential modeling.
The presence of LRD in input sequences poses a major computational challenge for sequential modeling in live production systems.
For tasks such as language understanding for auto-completion or behavioral prediction for recommendations, low latency deadlines (typically of the order of 100ms at most) have to be met by models. Such a tight constraint means that model architects need to find efficient ways to spend computational budgets in order to take information lying far in the past into account without incurring too much delay. The power-law decay in LRD sequences does imply a slower than expected decrease of the relevance of past information. It, however, also suggests 
that the most recent inputs are of higher relevance.
% However, it does also imply a significant decay which presents an opportunity in terms of model design.

\textbf{EvolutiveRNNs (EvoRNNs) for long LRD sequences.}
As information located far into the past is relevant but with a lower signal to noise ratio than the more recent observations, it is thus sub-optimal to spend the computational budget equally across inputs of different time steps as in standard RNNs. % less worthwhile spending our computational budget equally on the earliest inputs of the sequence.
Our insight is to spend our add/multiply budget at serving time in proportion of the amount of correlation between successive inputs and the final output.

We therefore propose an evolutive recurrent neural network architecture, named EvoRNN for LRD sequences. EvoRNN spends most of its computational budget on the most recent inputs while lowering the cost of considering inputs located further into the past. As looking further back into the past is less expensive for such models, practitioners can also consider longer sequences under the same latency constraints.

The main difference between the standard RNN equation~\ref{eq:standard_rnn} and the architecture we propose is that the size of the RNN cell being called depends on the distance to the end of the sequence:
\begin{equation}
    \label{eq:evo_rnn}
    [\hat{Y}_{t}, M_{t}]^T = \Phi_{L-t, \theta_{L - t}}\left(X_{t}, P_{L - t} M_{t-1}\right)
\end{equation}
where $L$ is the number of symbols in the sequence we consider, $t$ the current position and $P_{L - t}$ a rectangular matrix (whose parameters are learned) which re-sizes the previous state $M_{t-1}$ so that it has the dimension excepted for the state of cell $\Phi_{L-t, \theta_{L - t}}$.
In our experiments if the state dimensions are already compatible between the previous cell and the next cell we remove the projection matrix $P_{L - t}$.
We typically have few of these projections and plan to speed them up using fast structured linear operators such as randomized projections based on FFTs~\cite{yang2015deep}.

\begin{figure}[!ht]
    \centering
    \includegraphics[width=0.9\linewidth]{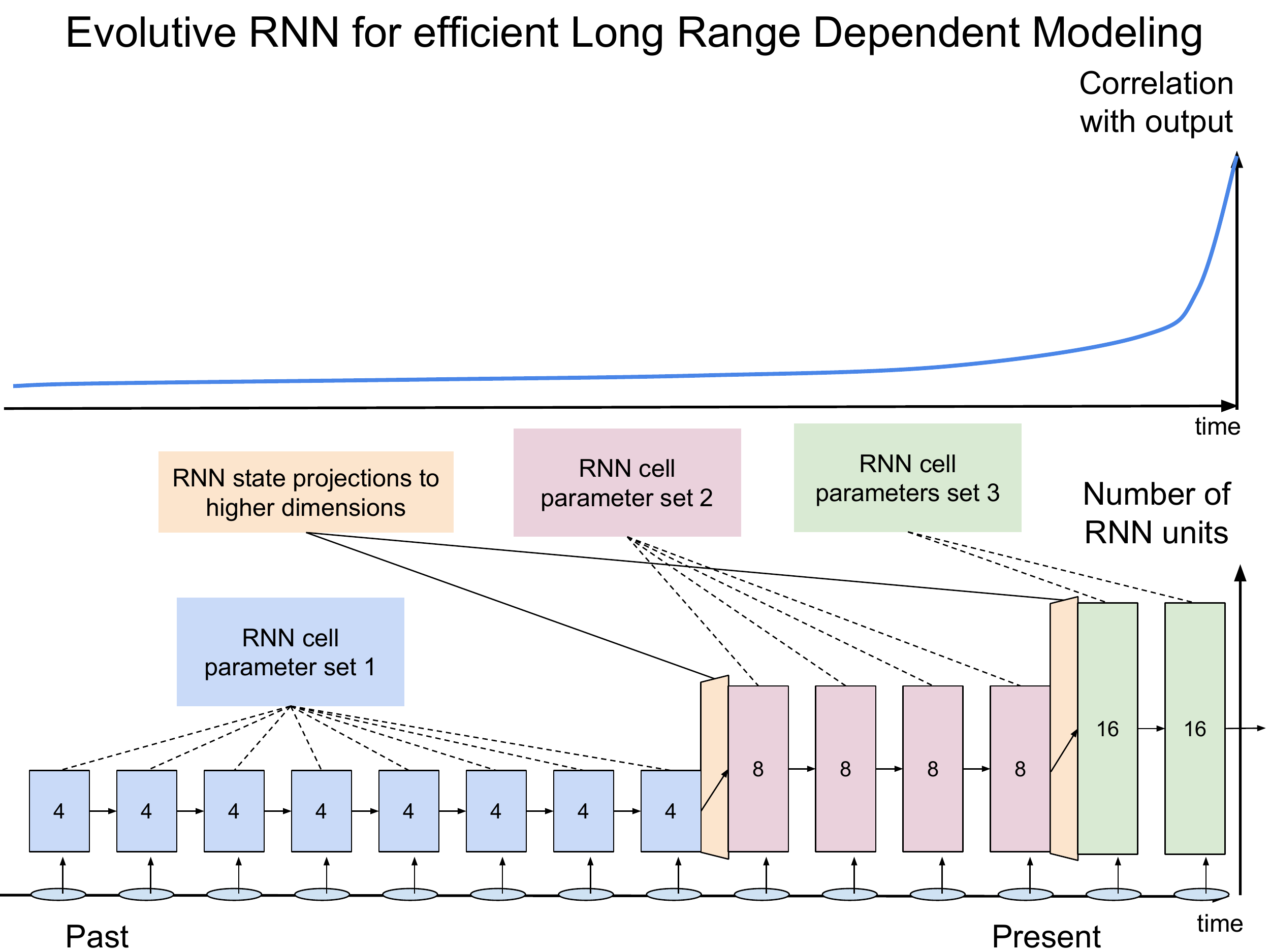}
    \caption{
\footnotesize{
    PowerLawEvoRNN architecture: the power-law decay of the computational intensity per input mirrors the power-law decay of correlation of inputs with the final output of the sequence in the presence of LRD.
}}
    \label{fig:evoRNN}
\end{figure}

\begin{figure}
    \centering
    \includegraphics[width=0.8\linewidth]{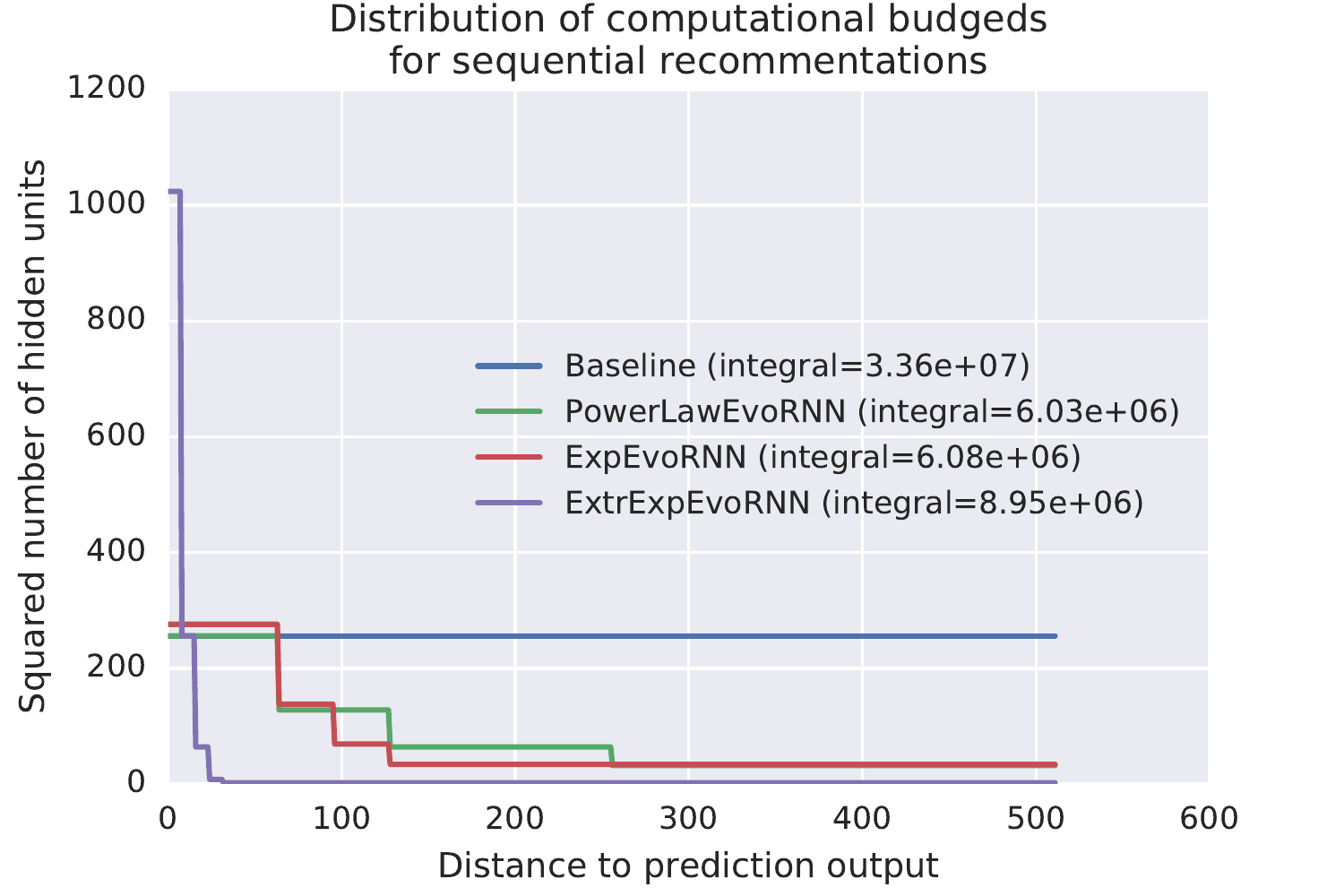}
    \caption{
\footnotesize{
    Computational budget -- $O(h^2)$ where $h$ is the number of hidden units -- of a baseline RNN and its PowerLawEvoRNN and ExpEvoRNN counterparts as a function of distance to end of sequence.
}}
    \label{fig:seq_rec_budgets}
\end{figure}

We consider two decay schemes for the number of units of the RNN cells as the distance to the end of the sequence -- where the prediction is made at serving time -- increases: a power-law decay (PowerLawEvoRNN) and an exponential decay (ExpEvoRNN). 
Figure~\ref{fig:evoRNN} illustrates our proposed architecture if a power-law decay of the computational intensity is chosen.
The computational budget of an RNN cell with $h$ hidden units scales as $O(h^2)$ to incorporate one input. The resulting distribution of compute time is presented in Figure~\ref{fig:seq_rec_budgets}.
One can appreciate how EvoRNN architectures rely on much fewer add/multiplies to process long sequences.

\section{Experimental results with EvoRNN}
We now compare the performance of EvoRNN architectures and standard RNNs on a benchmark language modeling task and a sequential recommendation task.
%after having presented measurements of the LRD coefficient in Section~\ref{sec:estimate}.
We show that on both tasks EvoRNNs outperform or reach comparable performance with the more computationally expensive baseline.
% We demonstrate on the sequential recommendation task -- which is clearly LRD -- that EvoRNNs perform as well as the more computationally expensive baseline.
% We show on the Billion word language modeling task -- which has weaker LRD -- that the EvoRNN architectures outperform the baseline.

\subsection{LM1B language modeling task}
The Billion word data set~\cite{chelba2013one} is a standard benchmark for language modeling~\cite{mikolov2010recurrent,bai2018empirical,chen2018dynamical} aimed at predicting the next word in a text.
We slightly modify the benchmark to create a long range prediction task involving longer sequences. 
Sequences of $128$ words are considered and the model's task is now to predict the last $4$ words.
% An architecture similar to the one shown in Figure~\ref{fig:estimation procedure} is employed for this predictive tasks and a vector embedding is learned for each word present in the dictionary. 

\begin{table*}
    \centering
    \begin{tabular}{cccr}
    \toprule
        \textbf{Model} & \textbf{Sub-sequence lengths} & \textbf{\# of hidden units} & \textbf{\# of add/mul.}\\
    \midrule
        LM baseline & $128$ & $2048$ & 536870912\\
    \midrule
        LM PowerLawEvoRNN & $64, 32, 16, 8, 4, 4$ & $64, 128, 256, 512, 1024, 2048$ & 24903680\\
    \midrule
        LM ExpEvoRNN & $108, 4, 4, 4, 4, 4$ & $64, 128, 256, 512, 1024, 2048$ & 22790144\\
    \midrule
        Seq. rec. baseline & $512$ & $256$ & $33554432$ \\
    \midrule
        Seq. rec. PowerLawEvoRNN & $256, 128, 64, 32, 32$ & $32, 64, 128, 256, 256$ & $6029312$\\
    \midrule
        Seq. rec. ExpEvoRNN & $384, 32, 32, 32, 32$ & $34, 69, 138, 276, 276$ & $6080928$\\
    \midrule
        Seq. rec. ExtrExpEvoRNN & $480, 8, 8, 8, 8$ & $2, 8, 64, 256, 1024$ & $8948096$\\
    \bottomrule
    \end{tabular}
    \caption{
    \footnotesize{
    RNN architectures employed in the sequential recommendation and language modeling tasks. Although they learn more parameters, EvoRNNs require much less compute time than baselines and therefore can be served under lower latency constraints. Here, \# add/mutiply give asymptotic complexity estimates to consider in relative magnitude.
}}
    \label{tab:num_units}
\end{table*}

The baseline model is a LSTM~\cite{hochreiter1997long}  following the implementation of~\cite{jozefowicz2016exploring}.  EvoRNNs follow the same setup except  the different distribution of compute resources. 
% We modified the baseline implementation of~\cite{jozefowicz2016exploring} and the sequence length is now $128$. 
% Again we employ the log-periodogram estimator and present the resulting estimated values of the LRD coefficient $d$ in Figure~\ref{fig:LRD_lm1b}.
% We observe that the amount of LRD is noticeably lower in this second dataset.
% In this sub-section, we now come back to the Billion Word language modeling benchmark and assess the performance of EvoRNN on this NLP task.
% We employ the EvoRNN architecture on a standard LSTM~\cite{hochreiter1997long} cell which is also used in the baseline. 
% An LSTM cell with input $(x_t)$ and output $(h_t)$ is defined by the following equations:
% \footnotesize{
% \begin{align}
% & \text{Forget gate:} & f_t = & \sigma(
% 	W^f_x x_t + W^f_h h_{t-1} + b^f
% ) \\
% & \text{Input gate:} & i_t = & \sigma(
% 	W^i_x x_t + W^i_h h_{t-1} + b^i
% ) \\
% & \text{Candidate:} &
% \tilde{c}_t = & \tanh(W_x x_t + W_h h_{t-1} + b)
% \label{eq:LSTM_candidate}
% \\
% & \text{Context update:} & 
% c_t = & f_t \odot \tilde{c}_t + i_t \odot c_{t-1}
% \label{eq:LSTM_ctx}
% \\
% & \text{Output gate:} &
% o_t = & \sigma
% \left(
% W^o_x x_t + W^o_h h_{t-1} + b^o
% \right)
% \\
% & \text{Output:} &
% h_t = &
% o_t \odot \tanh(c_t)
% \end{align}
% where
% $x_t \in \mathbb{R}^d$,
% $h_t \in \mathbb{R}^{d_h}$,
% $c_t \in \mathbb{R}^h$,
% $W_d \in \mathbb{R}^{d, d}$,
% $W_h \in \mathbb{R}^{d_h, d}$,
% $b \in \mathbb{R}^h$,
% $f_t \in \mathbb{R}^{d_h}$,
% $i_t \in \mathbb{R}^{d_h}$,
% $o_t \in \mathbb{R}^{d_h}$.
% }
We consider two variants of the EvoRNN architecture: a power law decay variant and an exponential decay variant.
The cell sizes and computational footprint of both variants are detailed in Table~\ref{tab:num_units}. As an example, the power law decay variant break the input sequence into six subsquences of length 64, 32, 16, 8, 4 and 4, each using RNNs of hidden units of 64, 128, 256, 512, 1024 and 2048, with most compute spending near the end of the sequence.~\footnote{
Note that as RNN cells of different number of hidden units are instantiated through the sequence, additional projection matrices, one between two sub-sequences, are learned in EvoRNNs. We can further save the computational cost by using fast random projections as in~\cite{yang2015deep}.} The total number of add/multiply performed using the baseline model as well as EvoRNNs with different scheduling are shown in the last column.

\begin{figure}[h!]
    \centering
    \includegraphics[width=0.9\linewidth]{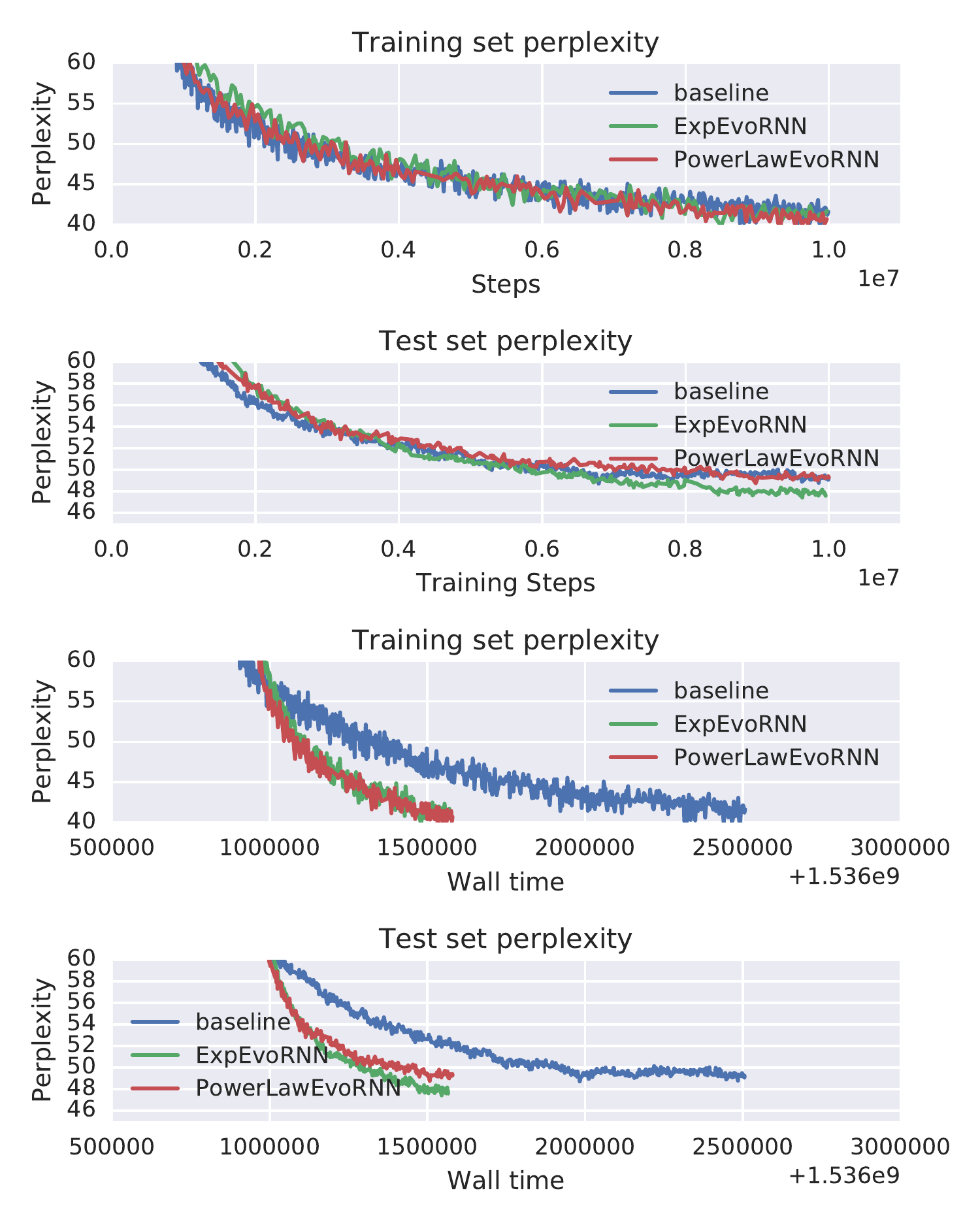}
    \caption{
\footnotesize{
    Performance results on language modeling task. EvoRNN architectures provide better performance than the baseline model and train considerably faster. The increase in training speed is expected as the EvoRNN architectures spend less compute-time on earlier inputs.
}}
    \label{fig:lm1b_task}
\end{figure}

Figure~\ref{fig:lm1b_task} shows that on this language modeling task both variants of the cheaper EvoRNN architectures out-perform the baseline model with only fraction of compute resources, as indicated by the wall time (bottom two plots) in addition to the estimate in Table~\ref{tab:num_units}. %  while the PowerLawRNN model is on par. 
This can be attributed to the implicit architectural prior of EvoRNN giving fewer degrees of freedom to parameters involved in processing inputs located further into the past, which inherently has less signal for predicting the next word near the end of the sequence. 
% Our estimates of $d$ in Figure~\ref{fig:LRD_lm1b} show a difference between word embeddings and interaction embeddings in Figure~\ref{fig:d_estimates_seq_rec} where components corresponding to the interface embedding had comparatively high values.
% Therefore one can argue that there is more memory in the sequential recommendation data-set than the language data-set.
Architectures having fewer hidden units for these inputs may resist more robustly to the lower levels of signal to noise ratio present at the beginning of the sequence.

\subsection{Sequential recommendation task}~\label{sec:experiment_recommendation}
% \begin{table}[t!]
% \center
% \caption{Sequential recommendation dataset.}\label{tb:seq_rec_dataset}
% \begin{tabular}{lc}
% \toprule
% \textbf{Production Dataset}\\
% \midrule
% Training sequences & >200M\\
% \midrule
% Test sequences & >1M\\
% \midrule
% Distinct items & $\approx$2M\\
% \midrule
% Avg. sequence length & $\sim200$\\
% \bottomrule
% \end{tabular}
% \end{table}

%\subsubsection{Symbol embeddings}
% The embeddings employed to represent the different items $v \in \mathcal{V}$ are trained as part of the sequential neural model aiming at predicting the next item the user will consume. 
% We now describe this RNN based model, it constitutes the baseline for our experiments on sequential recommendations.
Next we consider a sequential recommendation task, where the sequential recommender under consideration serves users accessing a
browsing page on which impressions are displayed. It has access to historical interactions, \textit{i.e.,} watched items, from the same user identified by the same personal account. 
% The impressions recommend items for users having previously used the website with the same personal account.
This sequential neural model nominating items for recommendation therefore maps the sequence of observations $\left\{\xi^u_{t_i} \right\}_{i = 1 \dots N_u}$ to a predicted item $\xi$. 
% Timestamps $t_i$ here mark the beginning of the consumption of an item.
% Here, each observation $\xi^u_{t_i}$ consists of several symbols it entails:
% a consumed item ID $v^u_{t_i}$, a creator ID $c^u_{t_i}$, a tag ID characterizing the page on which impressions are made (home vs item-specific page) $p^u_{t_i}$, and a software interface ID characterizing the OS employed by the user when browsing $s^u_{t_i}$.
% Key statistics of the dataset are summarized in Table~\ref{tb:seq_rec_dataset}.

Neural recommender systems attempt at foreseeing the interest of users under extreme constraints of latency and scale. We define the task as predicting the next item the user will consume given a recorded history of items already consumed. Such a problem setting is indeed common in collaborative filtering~\cite{sarwar2001item,linden2003amazon} recommendations. %We present here results obtained on a dataset where only about 700 000 pages are present that correspond to most popular pages. 
While the user history can span over months, only watches from the last 7 days are used for labels in training and watches in the last 2 days are used for testing.  The train/test split is $90/10\%$.
The test set does not overlap with the train set and corresponds to the last temporal slice of the dataset.
%The neural network predicts, for a sample of negatives, the probability that they are chosen and classically a negative sampling loss is employed in order to leverage observations belonging to a very large vocabulary~\cite{jean2014using}.
\begin{figure}
    \centering
    \includegraphics[trim=0.0cm 1cm 0.0cm 6cm, width=0.9\linewidth]{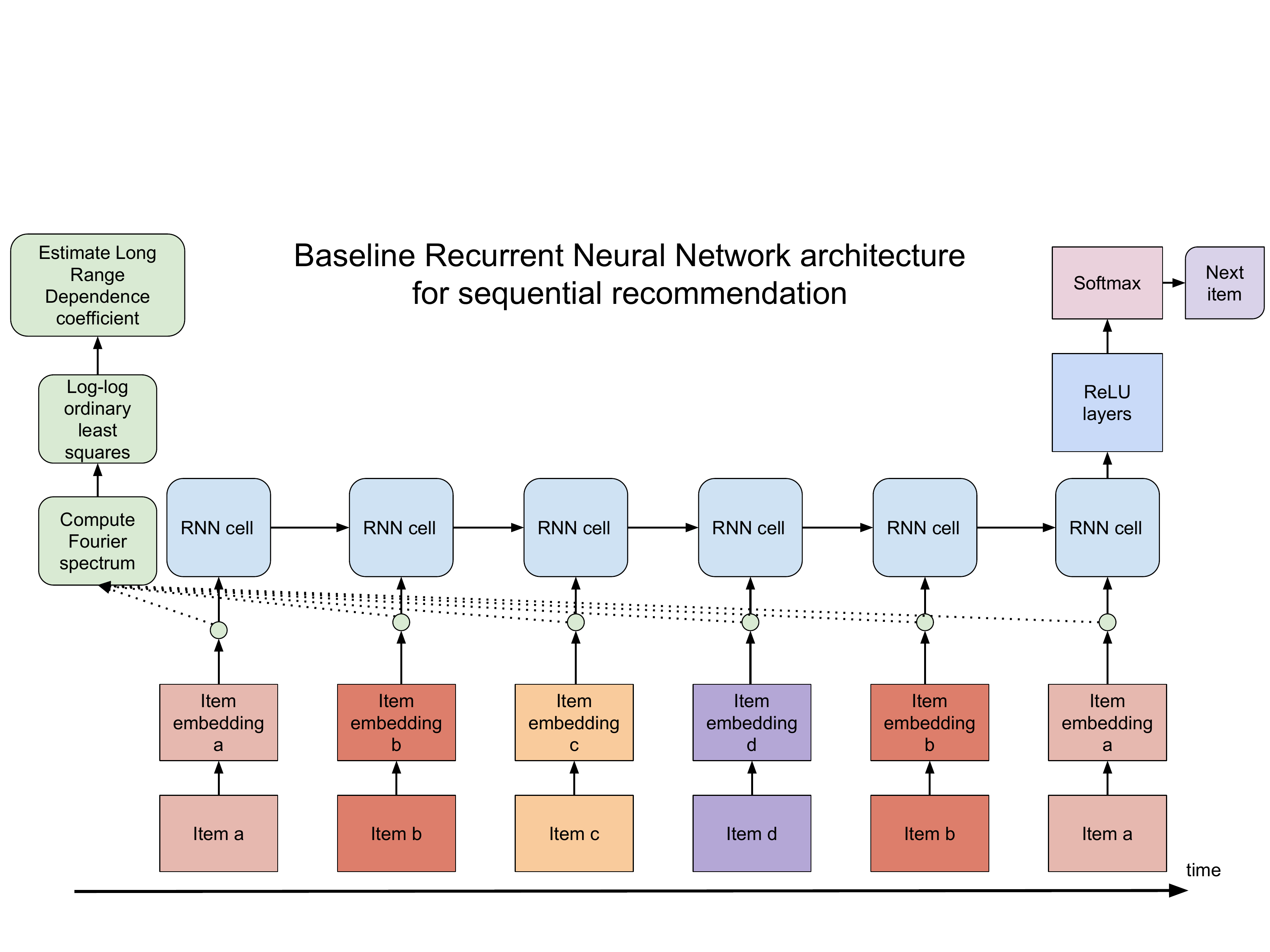}
    \caption{
\footnotesize{
Architecture for the sequential recommender. Input items are embedded into dense real-valued embeddings as described in section~\ref{sec:estimate_seq_rec} and sent through an RNN to predict the next item to be consumed. Left part of the figure, shown in green blocks, depicts the estimation process we detailed in section~\ref{sec:estimate_lm}
    % Representation of input items as dense real-valued embeddings and corresponding estimation procedure of the LRD coefficient of input sequences.
    % A sequential model is used to make predictions and symbol embeddings are trained as part of this model. 
    % The embeddings are analyzed by a log-periodogram estimator of the LRD coefficient $d$ before they are fed into the sequential model. The LRD coefficient measures the amount of memory in the input sequences which contrasts with pre-existing analysis procedures focusing on the ability of the model to memorize inputs.
    % The RNN architecture presented here is also the baseline that we modify later based on the insights revealed by the estimation of LRD coefficients.
    }
}
    \label{fig:estimation_procedure}
\end{figure}

Figure~\ref{fig:estimation_procedure} shows the underlying architecture powering the recommender. It relies on an RNN, a gated recurrent unit (GRU)~\cite{chung2014empirical}, to read through the sequence of past observations and predict the ID of the next item to be consumed by the user. The network parameters are trained using Adagrad to minimize a weighted cross-entropy loss.
The embedding dimensions and RNN cells are summarized in appendix (Table~\ref{tab:baseline_youtube_details}).
The model is a standard GRU fed with embedded symbols (after concatenation)
producing predictions on the items the user will click with a softmax layer trained by negative sampling.

Table~\ref{tab:num_units} shows the number of hidden units used in our baseline and the EvoRNNs for this task as well. Similarly, EvoRNNs only need a fraction of the add/multiplies used in the baseline RNN.

\begin{figure}
    \centering
    \includegraphics[width=0.75\linewidth]{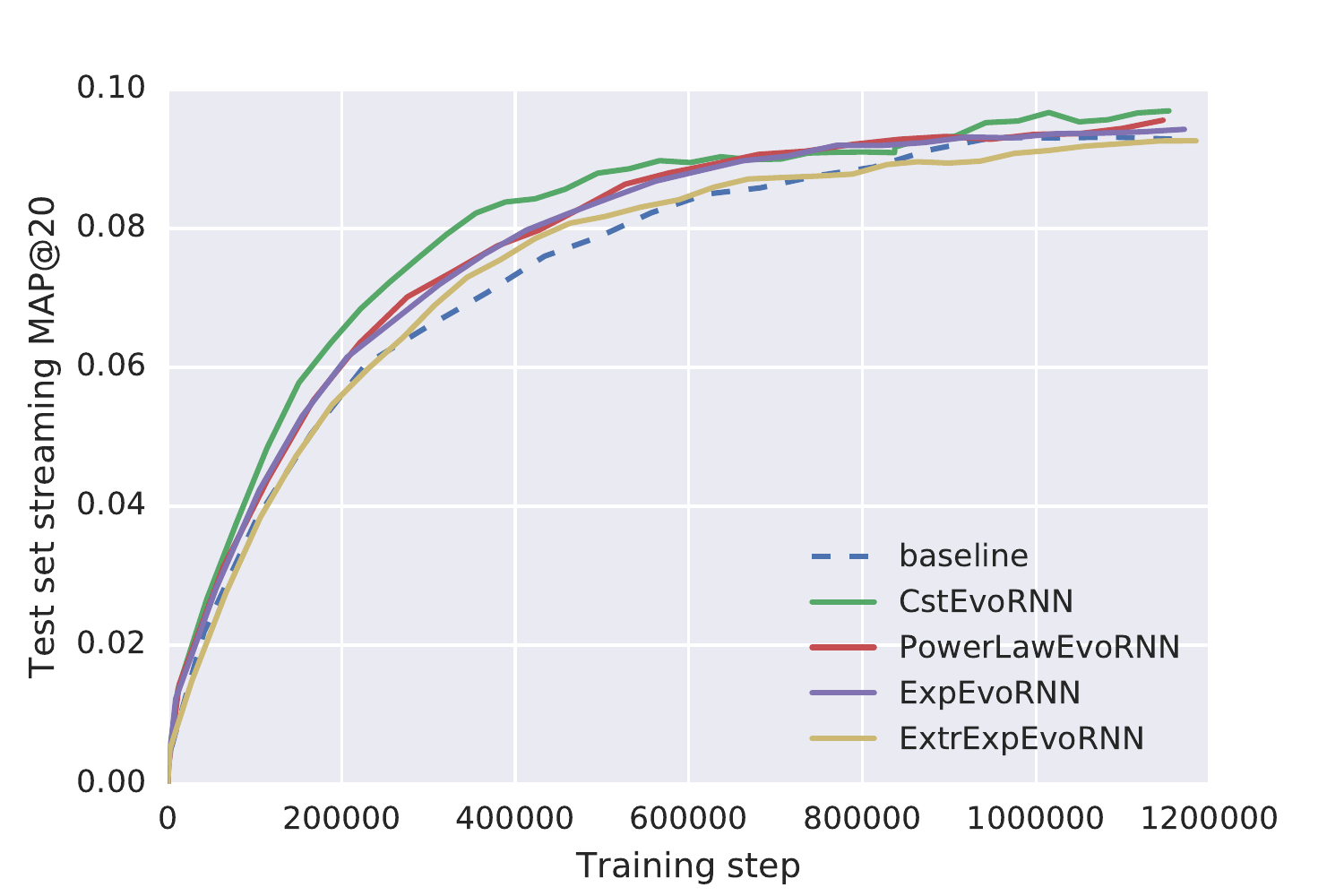}
    \caption{
\footnotesize{
    Performance results on the sequential recommendation task. For a lower computational budget, the EvoRNNs provide similar performance to the more expensive constant size baseline RNN. Although they use more memory because more parameters are learned in the present implementation, the PowerLaw and Exp EvoRNN networks need much fewer add/multiplies to compute a prediction as they spend less time processing inputs located further in the past. The extremely myopic ExtrExp RNN however does not perform as well as the baseline although it has many more free parameters and a higher computational budget than the other EvoRNNs.
    CstEvoRNN is an EvoRNN with as many independently learned cells as PowerLawEvoRNN but all as large as the baseline's cell. In spite of the increased number of free parameters, CstEvoRNN does not show much improvement.
}}
    \label{fig:seq_rec_task}
\end{figure}
We report the Mean-average-precision-at-20 (MAP@20)~\cite{guillaumin2009tagprop} as the main performance metric. Figure~\ref{fig:seq_rec_task} shows the progress of the MAP@20 score for the baseline and different variants of the EvoRNNs, up to $1.2$ million steps of training updates. We can see that 1) Variants of EvoRNNs reaches comparable performance as the baseline model, which uses 5 times more compute than EvoRNNs; 2) ExtrExpRNN, a myopic RNN designed to spend all of its compute near the very end of the sequences, and little on inputs in the past performed slightly worse, which confirms that the user behavior data is indeed LRD, and completely ignoring inputs from the past results in performance degradation.  % As training time is critical in order to prevent model staleness, we compare the MAP@20 of competing architectures as they reach $1.2$ million steps of training. 

\subsection{Conclusion and discussion on experiments}
After having demonstrated that the sequences of inputs we consider are LRD, we also showed that a parsimonious use of computational spending is sufficient to produce competitive expressive models for language modeling and sequential recommendation tasks.
The models we introduce, EvoRNNs, dedicate their serving time computational budget in priority to recent inputs while also leveraging information located further into the past. While the new architectures train more parameters (learning multiple RNN cells) in their present implementation, they help process longer sequences of inputs under a given latency deadline which is crucial to improve predictive accuracy for LRD sequential prediction tasks. In future work we aim at reducing the memory footprint by reusing parameters when processing earlier inputs in the sequence as larger parameter matrices can be projected to fewer dimensions.

\section{Conclusion}
While the issue of LRD has been considered widely in neural sequential modeling, LRD has not been quantified methodically for the corresponding sequences of inputs. For tasks such as language understanding and sequential recommendations --- where neural sequential models are pervasive and provide state-of-the-art performance --- model dependent gradient based considerations dominate when it comes to measuring LRD. 
In the present work, we employed a well established LRD theory for real-valued time series on sequences of vector-valued item embeddings to estimate LRD in sequences of discrete symbols belonging to large vocabularies. The resulting estimates of LRD coefficients unraveled new exploratory insights on modeling sequences of words and user interactions. Considering the power law decay of relevance of past inputs led to the construction of new recurrent architectures: EvoRNNs. EvoRNNs showed a performance at worst comparable with state-of-the-art baselines for language modeling and sequential recommendations using only a fraction of the computational cost.

\bibliographystyle{ACM-Reference-Format}
\bibliography{main}


\begin{thebibliography}{55}

%%% the Web address.

\ifx \showCODEN    \undefined \def \showCODEN     #1{\unskip}     \fi
\ifx \showDOI      \undefined \def \showDOI       #1{#1}\fi
\ifx \showISBNx    \undefined \def \showISBNx     #1{\unskip}     \fi
\ifx \showISBNxiii \undefined \def \showISBNxiii  #1{\unskip}     \fi
\ifx \showISSN     \undefined \def \showISSN      #1{\unskip}     \fi
\ifx \showLCCN     \undefined \def \showLCCN      #1{\unskip}     \fi
\ifx \shownote     \undefined \def \shownote      #1{#1}          \fi
\ifx \showarticletitle \undefined \def \showarticletitle #1{#1}   \fi
\ifx \showURL      \undefined \def \showURL       {\relax}        \fi
% invisible to TeX
\providecommand\bibfield[2]{#2}
\providecommand\bibinfo[2]{#2}
\providecommand\natexlab[1]{#1}
\providecommand\showeprint[2][]{arXiv:#2}

\bibitem[\protect\citeauthoryear{Arjovsky, Shah, and Bengio}{Arjovsky
  et~al\mbox{.}}{2016}]%
        {arjovsky2016unitary}
\bibfield{author}{\bibinfo{person}{Martin Arjovsky}, \bibinfo{person}{Amar
  Shah}, {and} \bibinfo{person}{Yoshua Bengio}.}
  \bibinfo{year}{2016}\natexlab{}.
\newblock \showarticletitle{Unitary evolution recurrent neural networks}. In
  \bibinfo{booktitle}{\emph{International Conference on Machine Learning}}.
  \bibinfo{pages}{1120--1128}.
\newblock


\bibitem[\protect\citeauthoryear{Bahdanau, Cho, and Bengio}{Bahdanau
  et~al\mbox{.}}{2014}]%
        {bahdanau2014neural}
\bibfield{author}{\bibinfo{person}{Dzmitry Bahdanau},
  \bibinfo{person}{Kyunghyun Cho}, {and} \bibinfo{person}{Yoshua Bengio}.}
  \bibinfo{year}{2014}\natexlab{}.
\newblock \showarticletitle{Neural machine translation by jointly learning to
  align and translate}.
\newblock \bibinfo{journal}{\emph{arXiv preprint arXiv:1409.0473}}
  (\bibinfo{year}{2014}).
\newblock


\bibitem[\protect\citeauthoryear{Bai, Kolter, and Koltun}{Bai
  et~al\mbox{.}}{2018}]%
        {bai2018empirical}
\bibfield{author}{\bibinfo{person}{Shaojie Bai}, \bibinfo{person}{J~Zico
  Kolter}, {and} \bibinfo{person}{Vladlen Koltun}.}
  \bibinfo{year}{2018}\natexlab{}.
\newblock \showarticletitle{An empirical evaluation of generic convolutional
  and recurrent networks for sequence modeling}.
\newblock \bibinfo{journal}{\emph{arXiv preprint arXiv:1803.01271}}
  (\bibinfo{year}{2018}).
\newblock


\bibitem[\protect\citeauthoryear{Belletti, Beutel, Jain, and Chi}{Belletti
  et~al\mbox{.}}{2018}]%
        {belletti2018factorized}
\bibfield{author}{\bibinfo{person}{Francois Belletti}, \bibinfo{person}{Alex
  Beutel}, \bibinfo{person}{Sagar Jain}, {and} \bibinfo{person}{Ed Chi}.}
  \bibinfo{year}{2018}\natexlab{}.
\newblock \showarticletitle{Factorized Recurrent Neural Architectures for
  Longer Range Dependence}. In \bibinfo{booktitle}{\emph{International
  Conference on Artificial Intelligence and Statistics}}.
  \bibinfo{pages}{1522--1530}.
\newblock


\bibitem[\protect\citeauthoryear{Belletti, Sparks, Bayen, and
  Gonzalez}{Belletti et~al\mbox{.}}{2017}]%
        {belletti2017random}
\bibfield{author}{\bibinfo{person}{Francois Belletti}, \bibinfo{person}{Evan
  Sparks}, \bibinfo{person}{Alexandre Bayen}, {and} \bibinfo{person}{Joseph
  Gonzalez}.} \bibinfo{year}{2017}\natexlab{}.
\newblock \showarticletitle{Random projection design for scalable implicit
  smoothing of randomly observed stochastic processes}. In
  \bibinfo{booktitle}{\emph{Artificial Intelligence and Statistics}}.
  \bibinfo{pages}{700--708}.
\newblock


\bibitem[\protect\citeauthoryear{Beran}{Beran}{2017}]%
        {beran2017statistics}
\bibfield{author}{\bibinfo{person}{Jan Beran}.}
  \bibinfo{year}{2017}\natexlab{}.
\newblock \bibinfo{booktitle}{\emph{Statistics for long-memory processes}}.
\newblock \bibinfo{publisher}{Routledge}.
\newblock


\bibitem[\protect\citeauthoryear{Brillinger}{Brillinger}{1981}]%
        {brillinger1981time}
\bibfield{author}{\bibinfo{person}{David~R Brillinger}.}
  \bibinfo{year}{1981}\natexlab{}.
\newblock \bibinfo{booktitle}{\emph{Time series: data analysis and theory}}.
  Vol.~\bibinfo{volume}{36}.
\newblock \bibinfo{publisher}{Siam}.
\newblock


\bibitem[\protect\citeauthoryear{Chang, Zhang, Han, Yu, Guo, Tan, Cui,
  Witbrock, Hasegawa-Johnson, and Huang}{Chang et~al\mbox{.}}{2017}]%
        {chang2017dilated}
\bibfield{author}{\bibinfo{person}{Shiyu Chang}, \bibinfo{person}{Yang Zhang},
  \bibinfo{person}{Wei Han}, \bibinfo{person}{Mo Yu}, \bibinfo{person}{Xiaoxiao
  Guo}, \bibinfo{person}{Wei Tan}, \bibinfo{person}{Xiaodong Cui},
  \bibinfo{person}{Michael Witbrock}, \bibinfo{person}{Mark~A
  Hasegawa-Johnson}, {and} \bibinfo{person}{Thomas~S Huang}.}
  \bibinfo{year}{2017}\natexlab{}.
\newblock \showarticletitle{Dilated recurrent neural networks}. In
  \bibinfo{booktitle}{\emph{Advances in Neural Information Processing
  Systems}}. \bibinfo{pages}{77--87}.
\newblock


\bibitem[\protect\citeauthoryear{Chelba, Mikolov, Schuster, Ge, Brants, Koehn,
  and Robinson}{Chelba et~al\mbox{.}}{2013}]%
        {chelba2013one}
\bibfield{author}{\bibinfo{person}{Ciprian Chelba}, \bibinfo{person}{Tomas
  Mikolov}, \bibinfo{person}{Mike Schuster}, \bibinfo{person}{Qi Ge},
  \bibinfo{person}{Thorsten Brants}, \bibinfo{person}{Phillipp Koehn}, {and}
  \bibinfo{person}{Tony Robinson}.} \bibinfo{year}{2013}\natexlab{}.
\newblock \showarticletitle{One billion word benchmark for measuring progress
  in statistical language modeling}.
\newblock \bibinfo{journal}{\emph{arXiv preprint arXiv:1312.3005}}
  (\bibinfo{year}{2013}).
\newblock


\bibitem[\protect\citeauthoryear{Chen, Pennington, and Schoenholz}{Chen
  et~al\mbox{.}}{2018}]%
        {chen2018dynamical}
\bibfield{author}{\bibinfo{person}{Minmin Chen}, \bibinfo{person}{Jeffrey
  Pennington}, {and} \bibinfo{person}{Samuel~S Schoenholz}.}
  \bibinfo{year}{2018}\natexlab{}.
\newblock \showarticletitle{Dynamical Isometry and a Mean Field Theory of RNNs:
  Gating Enables Signal Propagation in Recurrent Neural Networks}.
\newblock \bibinfo{journal}{\emph{arXiv preprint arXiv:1806.05394}}
  (\bibinfo{year}{2018}).
\newblock


\bibitem[\protect\citeauthoryear{Cho, Van~Merri{\"e}nboer, Gulcehre, Bahdanau,
  Bougares, Schwenk, and Bengio}{Cho et~al\mbox{.}}{2014}]%
        {cho2014learning}
\bibfield{author}{\bibinfo{person}{Kyunghyun Cho}, \bibinfo{person}{Bart
  Van~Merri{\"e}nboer}, \bibinfo{person}{Caglar Gulcehre},
  \bibinfo{person}{Dzmitry Bahdanau}, \bibinfo{person}{Fethi Bougares},
  \bibinfo{person}{Holger Schwenk}, {and} \bibinfo{person}{Yoshua Bengio}.}
  \bibinfo{year}{2014}\natexlab{}.
\newblock \showarticletitle{Learning phrase representations using RNN
  encoder-decoder for statistical machine translation}.
\newblock \bibinfo{journal}{\emph{arXiv preprint arXiv:1406.1078}}
  (\bibinfo{year}{2014}).
\newblock


\bibitem[\protect\citeauthoryear{Chung, Ahn, and Bengio}{Chung
  et~al\mbox{.}}{2016}]%
        {chung2016hierarchical}
\bibfield{author}{\bibinfo{person}{Junyoung Chung}, \bibinfo{person}{Sungjin
  Ahn}, {and} \bibinfo{person}{Yoshua Bengio}.}
  \bibinfo{year}{2016}\natexlab{}.
\newblock \showarticletitle{Hierarchical multiscale recurrent neural networks}.
\newblock \bibinfo{journal}{\emph{arXiv preprint arXiv:1609.01704}}
  (\bibinfo{year}{2016}).
\newblock


\bibitem[\protect\citeauthoryear{Chung, Gulcehre, Cho, and Bengio}{Chung
  et~al\mbox{.}}{2014}]%
        {chung2014empirical}
\bibfield{author}{\bibinfo{person}{Junyoung Chung}, \bibinfo{person}{Caglar
  Gulcehre}, \bibinfo{person}{KyungHyun Cho}, {and} \bibinfo{person}{Yoshua
  Bengio}.} \bibinfo{year}{2014}\natexlab{}.
\newblock \showarticletitle{Empirical evaluation of gated recurrent neural
  networks on sequence modeling}.
\newblock \bibinfo{journal}{\emph{arXiv preprint arXiv:1412.3555}}
  (\bibinfo{year}{2014}).
\newblock


\bibitem[\protect\citeauthoryear{Cover and Thomas}{Cover and Thomas}{2012}]%
        {cover2012elements}
\bibfield{author}{\bibinfo{person}{Thomas~M Cover} {and} \bibinfo{person}{Joy~A
  Thomas}.} \bibinfo{year}{2012}\natexlab{}.
\newblock \bibinfo{booktitle}{\emph{Elements of information theory}}.
\newblock \bibinfo{publisher}{John Wiley \& Sons}.
\newblock


\bibitem[\protect\citeauthoryear{Covington, Adams, and Sargin}{Covington
  et~al\mbox{.}}{2016}]%
        {covington2016deep}
\bibfield{author}{\bibinfo{person}{Paul Covington}, \bibinfo{person}{Jay
  Adams}, {and} \bibinfo{person}{Emre Sargin}.}
  \bibinfo{year}{2016}\natexlab{}.
\newblock \showarticletitle{Deep neural networks for youtube recommendations}.
  In \bibinfo{booktitle}{\emph{Proceedings of the 10th ACM Conference on
  Recommender Systems}}. ACM, \bibinfo{pages}{191--198}.
\newblock


\bibitem[\protect\citeauthoryear{Devooght and Bersini}{Devooght and
  Bersini}{2017}]%
        {devooght2017long}
\bibfield{author}{\bibinfo{person}{Robin Devooght} {and}
  \bibinfo{person}{Hugues Bersini}.} \bibinfo{year}{2017}\natexlab{}.
\newblock \showarticletitle{Long and short-term recommendations with recurrent
  neural networks}. In \bibinfo{booktitle}{\emph{Conference on User Modeling,
  Adaptation and Personalization}}. ACM, \bibinfo{pages}{13--21}.
\newblock


\bibitem[\protect\citeauthoryear{Doukhan, Oppenheim, and Taqqu}{Doukhan
  et~al\mbox{.}}{2002}]%
        {doukhan2002theory}
\bibfield{author}{\bibinfo{person}{Paul Doukhan}, \bibinfo{person}{George
  Oppenheim}, {and} \bibinfo{person}{Murad Taqqu}.}
  \bibinfo{year}{2002}\natexlab{}.
\newblock \bibinfo{booktitle}{\emph{Theory and applications of long-range
  dependence}}.
\newblock \bibinfo{publisher}{Springer Science \& Business Media}.
\newblock


\bibitem[\protect\citeauthoryear{Graves, Mohamed, and Hinton}{Graves
  et~al\mbox{.}}{2013}]%
        {graves2013speech}
\bibfield{author}{\bibinfo{person}{Alex Graves}, \bibinfo{person}{Abdel-rahman
  Mohamed}, {and} \bibinfo{person}{Geoffrey Hinton}.}
  \bibinfo{year}{2013}\natexlab{}.
\newblock \showarticletitle{Speech recognition with deep recurrent neural
  networks}. In \bibinfo{booktitle}{\emph{ICASSP}}. IEEE,
  \bibinfo{pages}{6645--6649}.
\newblock


\bibitem[\protect\citeauthoryear{Guillaumin, Mensink, Verbeek, and
  Schmid}{Guillaumin et~al\mbox{.}}{2009}]%
        {guillaumin2009tagprop}
\bibfield{author}{\bibinfo{person}{Matthieu Guillaumin},
  \bibinfo{person}{Thomas Mensink}, \bibinfo{person}{Jakob Verbeek}, {and}
  \bibinfo{person}{Cordelia Schmid}.} \bibinfo{year}{2009}\natexlab{}.
\newblock \showarticletitle{Tagprop: Discriminative metric learning in nearest
  neighbor models for image auto-annotation}. In
  \bibinfo{booktitle}{\emph{Computer Vision, 2009 IEEE 12th International
  Conference on}}. IEEE, \bibinfo{pages}{309--316}.
\newblock


\bibitem[\protect\citeauthoryear{Guo, Shamai, and Verd{\'u}}{Guo
  et~al\mbox{.}}{2005a}]%
        {guo2005additive}
\bibfield{author}{\bibinfo{person}{Dongning Guo}, \bibinfo{person}{Shlomo
  Shamai}, {and} \bibinfo{person}{Sergio Verd{\'u}}.}
  \bibinfo{year}{2005}\natexlab{a}.
\newblock \showarticletitle{Additive non-Gaussian noise channels: Mutual
  information and conditional mean estimation}. In
  \bibinfo{booktitle}{\emph{Information Theory, 2005. ISIT 2005. Proceedings.
  International Symposium on}}. IEEE, \bibinfo{pages}{719--723}.
\newblock


\bibitem[\protect\citeauthoryear{Guo, Shamai, and Verd{\'u}}{Guo
  et~al\mbox{.}}{2005b}]%
        {guo2005mutual}
\bibfield{author}{\bibinfo{person}{Dongning Guo}, \bibinfo{person}{Shlomo
  Shamai}, {and} \bibinfo{person}{Sergio Verd{\'u}}.}
  \bibinfo{year}{2005}\natexlab{b}.
\newblock \showarticletitle{Mutual information and minimum mean-square error in
  Gaussian channels}.
\newblock \bibinfo{journal}{\emph{IEEE Transactions on Information Theory}}
  \bibinfo{volume}{51}, \bibinfo{number}{4} (\bibinfo{year}{2005}),
  \bibinfo{pages}{1261--1282}.
\newblock


\bibitem[\protect\citeauthoryear{Hidasi, Karatzoglou, Baltrunas, and
  Tikk}{Hidasi et~al\mbox{.}}{2015}]%
        {hidasi2015session}
\bibfield{author}{\bibinfo{person}{Bal{\'a}zs Hidasi},
  \bibinfo{person}{Alexandros Karatzoglou}, \bibinfo{person}{Linas Baltrunas},
  {and} \bibinfo{person}{Domonkos Tikk}.} \bibinfo{year}{2015}\natexlab{}.
\newblock \showarticletitle{Session-based recommendations with recurrent neural
  networks}.
\newblock \bibinfo{journal}{\emph{arXiv preprint arXiv:1511.06939}}
  (\bibinfo{year}{2015}).
\newblock


\bibitem[\protect\citeauthoryear{Hochreiter and Schmidhuber}{Hochreiter and
  Schmidhuber}{1997}]%
        {hochreiter1997long}
\bibfield{author}{\bibinfo{person}{Sepp Hochreiter} {and}
  \bibinfo{person}{J{\"u}rgen Schmidhuber}.} \bibinfo{year}{1997}\natexlab{}.
\newblock \showarticletitle{Long short-term memory}.
\newblock \bibinfo{journal}{\emph{Neural computation}} \bibinfo{volume}{9},
  \bibinfo{number}{8} (\bibinfo{year}{1997}), \bibinfo{pages}{1735--1780}.
\newblock


\bibitem[\protect\citeauthoryear{Jing, Shen, Dub{\v{c}}ek, Peurifoy, Skirlo,
  LeCun, Tegmark, and Solja{\v{c}}i{\'c}}{Jing et~al\mbox{.}}{2016}]%
        {jing2016tunable}
\bibfield{author}{\bibinfo{person}{Li Jing}, \bibinfo{person}{Yichen Shen},
  \bibinfo{person}{Tena Dub{\v{c}}ek}, \bibinfo{person}{John Peurifoy},
  \bibinfo{person}{Scott Skirlo}, \bibinfo{person}{Yann LeCun},
  \bibinfo{person}{Max Tegmark}, {and} \bibinfo{person}{Marin
  Solja{\v{c}}i{\'c}}.} \bibinfo{year}{2016}\natexlab{}.
\newblock \showarticletitle{Tunable efficient unitary neural networks (EUNN)
  and their application to RNNs}.
\newblock \bibinfo{journal}{\emph{arXiv preprint arXiv:1612.05231}}
  (\bibinfo{year}{2016}).
\newblock


\bibitem[\protect\citeauthoryear{Jozefowicz, Vinyals, Schuster, Shazeer, and
  Wu}{Jozefowicz et~al\mbox{.}}{2016}]%
        {jozefowicz2016exploring}
\bibfield{author}{\bibinfo{person}{Rafal Jozefowicz}, \bibinfo{person}{Oriol
  Vinyals}, \bibinfo{person}{Mike Schuster}, \bibinfo{person}{Noam Shazeer},
  {and} \bibinfo{person}{Yonghui Wu}.} \bibinfo{year}{2016}\natexlab{}.
\newblock \showarticletitle{Exploring the limits of language modeling}.
\newblock \bibinfo{journal}{\emph{arXiv preprint arXiv:1602.02410}}
  (\bibinfo{year}{2016}).
\newblock


\bibitem[\protect\citeauthoryear{Lin and Tegmark}{Lin and Tegmark}{2016}]%
        {lin2016criticality}
\bibfield{author}{\bibinfo{person}{Henry~W Lin} {and} \bibinfo{person}{Max
  Tegmark}.} \bibinfo{year}{2016}\natexlab{}.
\newblock \showarticletitle{Criticality in formal languages and statistical
  physics}.
\newblock \bibinfo{journal}{\emph{arXiv preprint arXiv:1606.06737}}
  (\bibinfo{year}{2016}).
\newblock


\bibitem[\protect\citeauthoryear{Linden, Smith, and York}{Linden
  et~al\mbox{.}}{2003}]%
        {linden2003amazon}
\bibfield{author}{\bibinfo{person}{Greg Linden}, \bibinfo{person}{Brent Smith},
  {and} \bibinfo{person}{Jeremy York}.} \bibinfo{year}{2003}\natexlab{}.
\newblock \showarticletitle{Amazon. com recommendations: Item-to-item
  collaborative filtering}.
\newblock \bibinfo{journal}{\emph{IEEE Internet computing}}
  \bibinfo{volume}{7}, \bibinfo{number}{1} (\bibinfo{year}{2003}),
  \bibinfo{pages}{76--80}.
\newblock


\bibitem[\protect\citeauthoryear{Maaten and Hinton}{Maaten and Hinton}{2008}]%
        {maaten2008visualizing}
\bibfield{author}{\bibinfo{person}{Laurens van~der Maaten} {and}
  \bibinfo{person}{Geoffrey Hinton}.} \bibinfo{year}{2008}\natexlab{}.
\newblock \showarticletitle{Visualizing data using t-SNE}.
\newblock \bibinfo{journal}{\emph{Journal of machine learning research}}
  \bibinfo{volume}{9}, \bibinfo{number}{Nov} (\bibinfo{year}{2008}),
  \bibinfo{pages}{2579--2605}.
\newblock


\bibitem[\protect\citeauthoryear{MacKay}{MacKay}{2003}]%
        {mackay2003information}
\bibfield{author}{\bibinfo{person}{David~JC MacKay}.}
  \bibinfo{year}{2003}\natexlab{}.
\newblock \bibinfo{booktitle}{\emph{Information theory, inference and learning
  algorithms}}.
\newblock \bibinfo{publisher}{Cambridge university press}.
\newblock


\bibitem[\protect\citeauthoryear{Mandelbrot and Stewart}{Mandelbrot and
  Stewart}{1998}]%
        {mandelbrot1998fractals}
\bibfield{author}{\bibinfo{person}{Beno{\i}t~B Mandelbrot} {and}
  \bibinfo{person}{Ian Stewart}.} \bibinfo{year}{1998}\natexlab{}.
\newblock \showarticletitle{Fractals and scaling in finance}.
\newblock \bibinfo{journal}{\emph{Nature}} \bibinfo{volume}{391},
  \bibinfo{number}{6669} (\bibinfo{year}{1998}), \bibinfo{pages}{758--758}.
\newblock


\bibitem[\protect\citeauthoryear{Merity, Keskar, and Socher}{Merity
  et~al\mbox{.}}{2017}]%
        {merity2017regularizing}
\bibfield{author}{\bibinfo{person}{Stephen Merity},
  \bibinfo{person}{Nitish~Shirish Keskar}, {and} \bibinfo{person}{Richard
  Socher}.} \bibinfo{year}{2017}\natexlab{}.
\newblock \showarticletitle{Regularizing and optimizing LSTM language models}.
\newblock \bibinfo{journal}{\emph{arXiv preprint arXiv:1708.02182}}
  (\bibinfo{year}{2017}).
\newblock


\bibitem[\protect\citeauthoryear{Mikolov, Karafi{\'a}t, Burget,
  {\v{C}}ernock{\`y}, and Khudanpur}{Mikolov et~al\mbox{.}}{2010}]%
        {mikolov2010recurrent}
\bibfield{author}{\bibinfo{person}{Tom{\'a}{\v{s}} Mikolov},
  \bibinfo{person}{Martin Karafi{\'a}t}, \bibinfo{person}{Luk{\'a}{\v{s}}
  Burget}, \bibinfo{person}{Jan {\v{C}}ernock{\`y}}, {and}
  \bibinfo{person}{Sanjeev Khudanpur}.} \bibinfo{year}{2010}\natexlab{}.
\newblock \showarticletitle{Recurrent neural network based language model}. In
  \bibinfo{booktitle}{\emph{Eleventh Annual Conference of the International
  Speech Communication Association}}.
\newblock


\bibitem[\protect\citeauthoryear{Mikolov, Sutskever, Chen, Corrado, and
  Dean}{Mikolov et~al\mbox{.}}{2013a}]%
        {mikolov2013distributed}
\bibfield{author}{\bibinfo{person}{Tomas Mikolov}, \bibinfo{person}{Ilya
  Sutskever}, \bibinfo{person}{Kai Chen}, \bibinfo{person}{Greg~S Corrado},
  {and} \bibinfo{person}{Jeff Dean}.} \bibinfo{year}{2013}\natexlab{a}.
\newblock \showarticletitle{Distributed representations of words and phrases
  and their compositionality}. In \bibinfo{booktitle}{\emph{Advances in neural
  information processing systems}}. \bibinfo{pages}{3111--3119}.
\newblock


\bibitem[\protect\citeauthoryear{Mikolov, Yih, and Zweig}{Mikolov
  et~al\mbox{.}}{2013b}]%
        {mikolov2013linguistic}
\bibfield{author}{\bibinfo{person}{Tomas Mikolov}, \bibinfo{person}{Wen-tau
  Yih}, {and} \bibinfo{person}{Geoffrey Zweig}.}
  \bibinfo{year}{2013}\natexlab{b}.
\newblock \showarticletitle{Linguistic regularities in continuous space word
  representations}. In \bibinfo{booktitle}{\emph{Proceedings of the 2013
  Conference of the North American Chapter of the Association for Computational
  Linguistics: Human Language Technologies}}. \bibinfo{pages}{746--751}.
\newblock


\bibitem[\protect\citeauthoryear{Miller and Hardt}{Miller and Hardt}{2018}]%
        {miller2018recurrent}
\bibfield{author}{\bibinfo{person}{John Miller} {and} \bibinfo{person}{Moritz
  Hardt}.} \bibinfo{year}{2018}\natexlab{}.
\newblock \showarticletitle{When Recurrent Models Don't Need To Be Recurrent}.
\newblock \bibinfo{journal}{\emph{arXiv preprint arXiv:1805.10369}}
  (\bibinfo{year}{2018}).
\newblock


\bibitem[\protect\citeauthoryear{Pascanu, Mikolov, and Bengio}{Pascanu
  et~al\mbox{.}}{2013}]%
        {pascanu2013difficulty}
\bibfield{author}{\bibinfo{person}{Razvan Pascanu}, \bibinfo{person}{Tomas
  Mikolov}, {and} \bibinfo{person}{Yoshua Bengio}.}
  \bibinfo{year}{2013}\natexlab{}.
\newblock \showarticletitle{On the difficulty of training recurrent neural
  networks}. In \bibinfo{booktitle}{\emph{International Conference on Machine
  Learning}}. \bibinfo{pages}{1310--1318}.
\newblock


\bibitem[\protect\citeauthoryear{Pennington, Socher, and Manning}{Pennington
  et~al\mbox{.}}{2014}]%
        {pennington2014glove}
\bibfield{author}{\bibinfo{person}{Jeffrey Pennington},
  \bibinfo{person}{Richard Socher}, {and} \bibinfo{person}{Christopher
  Manning}.} \bibinfo{year}{2014}\natexlab{}.
\newblock \showarticletitle{Glove: Global vectors for word representation}. In
  \bibinfo{booktitle}{\emph{Proceedings of the 2014 conference on empirical
  methods in natural language processing (EMNLP)}}.
  \bibinfo{pages}{1532--1543}.
\newblock


\bibitem[\protect\citeauthoryear{Pipiras and Taqqu}{Pipiras and Taqqu}{2017}]%
        {pipiras2017long}
\bibfield{author}{\bibinfo{person}{Vladas Pipiras} {and}
  \bibinfo{person}{Murad~S Taqqu}.} \bibinfo{year}{2017}\natexlab{}.
\newblock \bibinfo{booktitle}{\emph{Long-range dependence and
  self-similarity}}. Vol.~\bibinfo{volume}{45}.
\newblock \bibinfo{publisher}{Cambridge university press}.
\newblock


\bibitem[\protect\citeauthoryear{Quadrana, Karatzoglou, Hidasi, and
  Cremonesi}{Quadrana et~al\mbox{.}}{2017}]%
        {quadrana2017personalizing}
\bibfield{author}{\bibinfo{person}{Massimo Quadrana},
  \bibinfo{person}{Alexandros Karatzoglou}, \bibinfo{person}{Bal{\'a}zs
  Hidasi}, {and} \bibinfo{person}{Paolo Cremonesi}.}
  \bibinfo{year}{2017}\natexlab{}.
\newblock \showarticletitle{Personalizing session-based recommendations with
  hierarchical recurrent neural networks}. In
  \bibinfo{booktitle}{\emph{Proceedings of the Eleventh ACM Conference on
  Recommender Systems}}. ACM, \bibinfo{pages}{130--137}.
\newblock


\bibitem[\protect\citeauthoryear{Rahimi and Recht}{Rahimi and Recht}{2008}]%
        {rahimi2008random}
\bibfield{author}{\bibinfo{person}{Ali Rahimi} {and} \bibinfo{person}{Benjamin
  Recht}.} \bibinfo{year}{2008}\natexlab{}.
\newblock \showarticletitle{Random features for large-scale kernel machines}.
  In \bibinfo{booktitle}{\emph{Advances in neural information processing
  systems}}. \bibinfo{pages}{1177--1184}.
\newblock


\bibitem[\protect\citeauthoryear{Robinson}{Robinson}{1995}]%
        {robinson1995log}
\bibfield{author}{\bibinfo{person}{Peter~M Robinson}.}
  \bibinfo{year}{1995}\natexlab{}.
\newblock \showarticletitle{Log-periodogram regression of time series with long
  range dependence}.
\newblock \bibinfo{journal}{\emph{The annals of Statistics}}
  (\bibinfo{year}{1995}), \bibinfo{pages}{1048--1072}.
\newblock


\bibitem[\protect\citeauthoryear{Samorodnitsky et~al\mbox{.}}{Samorodnitsky
  et~al\mbox{.}}{2007}]%
        {samorodnitsky2007long}
\bibfield{author}{\bibinfo{person}{Gennady Samorodnitsky} {et~al\mbox{.}}}
  \bibinfo{year}{2007}\natexlab{}.
\newblock \showarticletitle{Long range dependence}.
\newblock \bibinfo{journal}{\emph{Foundations and Trends{\textregistered} in
  Stochastic Systems}} \bibinfo{volume}{1}, \bibinfo{number}{3}
  (\bibinfo{year}{2007}), \bibinfo{pages}{163--257}.
\newblock


\bibitem[\protect\citeauthoryear{Sarwar, Karypis, Konstan, and Riedl}{Sarwar
  et~al\mbox{.}}{2001}]%
        {sarwar2001item}
\bibfield{author}{\bibinfo{person}{Badrul Sarwar}, \bibinfo{person}{George
  Karypis}, \bibinfo{person}{Joseph Konstan}, {and} \bibinfo{person}{John
  Riedl}.} \bibinfo{year}{2001}\natexlab{}.
\newblock \showarticletitle{Item-based collaborative filtering recommendation
  algorithms}. In \bibinfo{booktitle}{\emph{Proceedings of the 10th
  international conference on World Wide Web}}. ACM, \bibinfo{pages}{285--295}.
\newblock


\bibitem[\protect\citeauthoryear{Smirnova and Vasile}{Smirnova and
  Vasile}{2017}]%
        {smirnova2017contextual}
\bibfield{author}{\bibinfo{person}{Elena Smirnova} {and}
  \bibinfo{person}{Flavian Vasile}.} \bibinfo{year}{2017}\natexlab{}.
\newblock \showarticletitle{Contextual sequence modeling for recommendation
  with recurrent neural networks}. In \bibinfo{booktitle}{\emph{Proceedings of
  the 2nd Workshop on Deep Learning for Recommender Systems}}. ACM,
  \bibinfo{pages}{2--9}.
\newblock


\bibitem[\protect\citeauthoryear{Sornette}{Sornette}{2006}]%
        {sornette2006critical}
\bibfield{author}{\bibinfo{person}{Didier Sornette}.}
  \bibinfo{year}{2006}\natexlab{}.
\newblock \bibinfo{booktitle}{\emph{Critical phenomena in natural sciences:
  chaos, fractals, selforganization and disorder: concepts and tools}}.
\newblock \bibinfo{publisher}{Springer Science \& Business Media}.
\newblock


\bibitem[\protect\citeauthoryear{Sutskever, Vinyals, and Le}{Sutskever
  et~al\mbox{.}}{2014}]%
        {sutskever2014sequence}
\bibfield{author}{\bibinfo{person}{Ilya Sutskever}, \bibinfo{person}{Oriol
  Vinyals}, {and} \bibinfo{person}{Quoc~V Le}.}
  \bibinfo{year}{2014}\natexlab{}.
\newblock \showarticletitle{Sequence to sequence learning with neural
  networks}. In \bibinfo{booktitle}{\emph{Advances in neural information
  processing systems}}. \bibinfo{pages}{3104--3112}.
\newblock


\bibitem[\protect\citeauthoryear{Tang, Belletti, Jain, Chen, Beutel, Xu, and
  Chi}{Tang et~al\mbox{.}}{2019}]%
        {tang2019towards}
\bibfield{author}{\bibinfo{person}{Jiaxi Tang}, \bibinfo{person}{Francois
  Belletti}, \bibinfo{person}{Sagar Jain}, \bibinfo{person}{Minmin Chen},
  \bibinfo{person}{Alex Beutel}, \bibinfo{person}{Can Xu}, {and}
  \bibinfo{person}{Ed~H Chi}.} \bibinfo{year}{2019}\natexlab{}.
\newblock \showarticletitle{Towards Neural Mixture Recommender for Long Range
  Dependent User Sequences}.
\newblock \bibinfo{journal}{\emph{arXiv preprint arXiv:1902.08588}}
  (\bibinfo{year}{2019}).
\newblock


\bibitem[\protect\citeauthoryear{Trinh, Dai, Luong, and Le}{Trinh
  et~al\mbox{.}}{2018}]%
        {trinh2018learning}
\bibfield{author}{\bibinfo{person}{Trieu~H Trinh}, \bibinfo{person}{Andrew~M
  Dai}, \bibinfo{person}{Thang Luong}, {and} \bibinfo{person}{Quoc~V Le}.}
  \bibinfo{year}{2018}\natexlab{}.
\newblock \showarticletitle{Learning longer-term dependencies in rnns with
  auxiliary losses}.
\newblock \bibinfo{journal}{\emph{arXiv preprint arXiv:1803.00144}}
  (\bibinfo{year}{2018}).
\newblock


\bibitem[\protect\citeauthoryear{Van Den~Oord, Dieleman, Zen, Simonyan,
  Vinyals, Graves, Kalchbrenner, Senior, and Kavukcuoglu}{Van Den~Oord
  et~al\mbox{.}}{2016}]%
        {van2016wavenet}
\bibfield{author}{\bibinfo{person}{A{\"a}ron Van Den~Oord},
  \bibinfo{person}{Sander Dieleman}, \bibinfo{person}{Heiga Zen},
  \bibinfo{person}{Karen Simonyan}, \bibinfo{person}{Oriol Vinyals},
  \bibinfo{person}{Alex Graves}, \bibinfo{person}{Nal Kalchbrenner},
  \bibinfo{person}{Andrew~W Senior}, {and} \bibinfo{person}{Koray
  Kavukcuoglu}.} \bibinfo{year}{2016}\natexlab{}.
\newblock \showarticletitle{WaveNet: A generative model for raw audio.}. In
  \bibinfo{booktitle}{\emph{SSW}}. \bibinfo{pages}{125}.
\newblock


\bibitem[\protect\citeauthoryear{Vaswani, Shazeer, Parmar, Uszkoreit, Jones,
  Gomez, Kaiser, and Polosukhin}{Vaswani et~al\mbox{.}}{2017}]%
        {vaswani2017attention}
\bibfield{author}{\bibinfo{person}{Ashish Vaswani}, \bibinfo{person}{Noam
  Shazeer}, \bibinfo{person}{Niki Parmar}, \bibinfo{person}{Jakob Uszkoreit},
  \bibinfo{person}{Llion Jones}, \bibinfo{person}{Aidan~N Gomez},
  \bibinfo{person}{{\L}ukasz Kaiser}, {and} \bibinfo{person}{Illia
  Polosukhin}.} \bibinfo{year}{2017}\natexlab{}.
\newblock \showarticletitle{Attention is all you need}. In
  \bibinfo{booktitle}{\emph{Advances in neural information processing
  systems}}. \bibinfo{pages}{5998--6008}.
\newblock


\bibitem[\protect\citeauthoryear{Vorontsov, Trabelsi, Kadoury, and
  Pal}{Vorontsov et~al\mbox{.}}{2017}]%
        {vorontsov2017orthogonality}
\bibfield{author}{\bibinfo{person}{Eugene Vorontsov}, \bibinfo{person}{Chiheb
  Trabelsi}, \bibinfo{person}{Samuel Kadoury}, {and} \bibinfo{person}{Chris
  Pal}.} \bibinfo{year}{2017}\natexlab{}.
\newblock \showarticletitle{On orthogonality and learning recurrent networks
  with long term dependencies}.
\newblock \bibinfo{journal}{\emph{arXiv preprint arXiv:1702.00071}}
  (\bibinfo{year}{2017}).
\newblock


\bibitem[\protect\citeauthoryear{Wu, Ahmed, Beutel, Smola, and Jing}{Wu
  et~al\mbox{.}}{2017}]%
        {wu2017recurrent}
\bibfield{author}{\bibinfo{person}{Chao-Yuan Wu}, \bibinfo{person}{Amr Ahmed},
  \bibinfo{person}{Alex Beutel}, \bibinfo{person}{Alexander~J Smola}, {and}
  \bibinfo{person}{How Jing}.} \bibinfo{year}{2017}\natexlab{}.
\newblock \showarticletitle{Recurrent recommender networks}. In
  \bibinfo{booktitle}{\emph{Proceedings of the tenth ACM international
  conference on web search and data mining}}. ACM, \bibinfo{pages}{495--503}.
\newblock


\bibitem[\protect\citeauthoryear{Xin, Mayoraz, Pham, Lakshmanan, and
  Anderson}{Xin et~al\mbox{.}}{2017}]%
        {xin2017folding}
\bibfield{author}{\bibinfo{person}{Doris Xin}, \bibinfo{person}{Nicolas
  Mayoraz}, \bibinfo{person}{Hubert Pham}, \bibinfo{person}{Karthik
  Lakshmanan}, {and} \bibinfo{person}{John~R Anderson}.}
  \bibinfo{year}{2017}\natexlab{}.
\newblock \showarticletitle{Folding: Why Good Models Sometimes Make Spurious
  Recommendations}. In \bibinfo{booktitle}{\emph{Proceedings of the Eleventh
  ACM Conference on Recommender Systems}}. ACM, \bibinfo{pages}{201--209}.
\newblock


\bibitem[\protect\citeauthoryear{Yang, Moczulski, Denil, de~Freitas, Smola,
  Song, and Wang}{Yang et~al\mbox{.}}{2015}]%
        {yang2015deep}
\bibfield{author}{\bibinfo{person}{Zichao Yang}, \bibinfo{person}{Marcin
  Moczulski}, \bibinfo{person}{Misha Denil}, \bibinfo{person}{Nando de
  Freitas}, \bibinfo{person}{Alex Smola}, \bibinfo{person}{Le Song}, {and}
  \bibinfo{person}{Ziyu Wang}.} \bibinfo{year}{2015}\natexlab{}.
\newblock \showarticletitle{Deep fried convnets}. In
  \bibinfo{booktitle}{\emph{Proceedings of the IEEE International Conference on
  Computer Vision}}. \bibinfo{pages}{1476--1483}.
\newblock


\bibitem[\protect\citeauthoryear{Yu and Koltun}{Yu and Koltun}{2015}]%
        {yu2015multi}
\bibfield{author}{\bibinfo{person}{Fisher Yu} {and} \bibinfo{person}{Vladlen
  Koltun}.} \bibinfo{year}{2015}\natexlab{}.
\newblock \showarticletitle{Multi-scale context aggregation by dilated
  convolutions}.
\newblock \bibinfo{journal}{\emph{arXiv preprint arXiv:1511.07122}}
  (\bibinfo{year}{2015}).
\newblock


\end{thebibliography}

\newpage

\section*{Appendix}
We now produce the pseudo code and instructions to help reproduce our experiments.

\subsection*{Implementation of estimators for memory coefficients $d \in \mathbb{R}^p$ on large data sets}
We only described in Algorithm~\ref{alg:OLS_procedure} how we implemented the log-periodogram method to estimate LRD coefficients for a single sequence of symbols.
We now give more details on how the estimate at the scale of an entire data set comprising millions of such sequences is computed.
We employ an exponential moving average based estimation as follows.
Consider embeddings of $p$ dimensions, the LRD coefficient vector we estimate is therefore in $\mathbb{R}^d$.
We start with a first estimate $\widehat{d}_0 \in \mathbb{R}^p$ and chose a learning rate sequence $(\alpha_i)_{i \in \mathbb{N}}$.
Sequences are grouped in mini-batches (typically of $256$ sequences) on which log-periodogram estimates are computed independently (in parallel) by running Algorithm~\ref{alg:OLS_procedure}.
Considering a batch size $B$ and the step $i$ of the SGD, let $(\widetilde{d}_{i,j})_{j = 0, \dots, B - 1}$ the resulting batch of LRD coefficient estimates in $\mathbb{R}^p$.
The previous estimate for the overall LRD coefficient is updated as follows:
$$
    \widehat{d}_{i+1} = (1 - \alpha_i) \widehat{d}_{i} + \alpha_i \frac{1}{B} \sum_{j = 0}^{B - 1} \widetilde{d}_{i,j}.
$$
In our experiments the learning rate follows a standard inverse decay scheme ($\sim \frac{1}{i}$).

\subsection*{Implementation of Neural architectures}
The appendix now presents the general neural architecture for symbol prediction considered in the paper before describing the particular implementations for language understanding and behavior prediction.

\subsubsection*{Generic RNN for symbol prediction}
We now describe the generic architecture employed in the paper to predict the next symbol in a sequence of symbols.
Let $\text{Vocab} = \left\{0 \dots V - 1\right\}$ be the vocabulary of possible symbol values.
After having chosen an input embedding size $d^{\text{input}}$ we create a table of input embeddings:
$$
    E^{\text{input}} = (e^{\text{input}}_i)_{i = 1 \dots V - 1}
$$
where $\forall i \in \left\{1 \dots V - 1\right\}, e^{\text{input}}_i \in \mathbb{R}^{d^{\text{input}}}$.

Consider a sequence of input symbols $(s_0, \dots, s_{L-1})$. We attempt to predict the next symbol $s_L$, i.e. the label.
The very first processing step transforms the sequence of $L$ symbols in $\text{Vocab}$ into a sequence of real valued vectors:
$$
    (s_0, \dots, s_{L-1}) \rightarrow (e^{\text{input}}_{s_0}, \ldots, e^{\text{input}}_{s_{L-1}}).
$$

Now, $(e^{\text{input}}_{s_0}, \ldots, e^{\text{input}}_{s_{L-1}})$ constitutes the sequence that serves as input into the first (and possibly only) RNN layer: $\text{RNN}_0$.
Let us describe how the $i^{\text{th}}$ RNN layer processes its inputs denoted by $(x^i_0, \dots, x^i_{L-1})$.

Consider an RNN layer taking inputs in $\mathbb{R}^{d^i}$ and outputting an RNN state in
$\mathbb{R}^{d^{i + 1}}$ as well as an output in $\mathbb{R}^{d^{i + 1}}$:
$$
    (x^i_0, \ldots, x^i_{L-1}) \rightarrow (x^{i + 1}_0, \dots, x^{i + 1}_{L-1})
$$
where the outputs 
$(x^{i + 1}_0, \dots, o^{i + 1}_{L-1})$ 
and the states 
$(h^i_0, \dots, h^i_{L-1})$
are generated --- given an initial value $\text{init}^i \in \mathbb{R}^{d^{i + 1}}$ for the RNN state which can be set to $0$, randomized or learned as a parameter --- by the recurrent relation:
\begin{align}
    & h^{i}_0 = \text{init}^i, \\
    & (x^{i + 1}_{t + 1}, h^i_{t + 1}) = 
    \text{RNNCell}_{\theta^i}^i (x^i_t, h^i_t).
    \label{eq:appendix_rec}
\end{align}

Typically, $\text{RNNCell}_{\theta}^i (x^i_t, h^i_t)$ (if RNNCell is an Elman/Jordan cell, an LSTM cell, a GRU cell for instance) involves $O(d^{i} \times d^{i + 1})$ storage for learned parameters and $O(d^{i} \times d^{i + 1})$ add/multiplies because of the underlying matrix/vector multiplications.

For an RNN network of depth $D$, i.e. with $D$ stacked RNN layers $\text{RNN}^0, \dots, \text{RNN}^{D - 1}$, the final output of interest is
$$
    (x^D_L).
$$
One may also predict further symbols such as $s_{L + 1}, s_{L + 2}, \dots, s_{L + H}$.
This is readily possible by unfolding the RNN recurrence further in time after the time steps $t$.
In that case, the RNN outputs of interest are
$$
    (x^D_L, x^D_{L + 1}, \dots, x^D_{L + H}).
$$

In either case, a decoding layer (which may in fact comprise several Fully Connected layers) transforms the RNN outputs into the final outputs used for symbol prediction.
Let $\text{Dec}_{\theta^{\text{Dec}}}$ be the final layer taking inputs in 
$\mathbb{R}^{d^D}$ and with outputs in $\mathbb{R}^{d^O}$.
The decoded output sequence is obtained as follows:
$$
    (x^D_L, x^D_{L + 1}, \dots, x^D_{L + H}) \rightarrow
    (y_L, y_{L + 1}, \dots, y_{L + H})
$$
where
$$
    (y_L, y_{L + 1}, \dots, y_{L + H}) =
    (\text{Dec}_{\theta^{\text{Dec}}}(x^D_L), \dots, \text{Dec}_{\theta^{\text{Dec}}}(x^D_{L + H})).
$$

Finally, to match the vector valued outputs with the labels, we classically employ 
an output side embedding table 
$E^{\text{output}}$
(which may be in fact the input side embedding table $E^{\text{input}}$ itself if the corresponding embedding dimensions are equal):
$$
    E^{\text{output}} = (e^{\text{output}}_i)_{i = 1 \dots V - 1}
$$
where $\forall i \in \left\{1 \dots V - 1\right\}, e^{\text{output}}_i \in \mathbb{R}^{d^{\text{output}}}$.

The output embedding table encodes the sequence of symbols that have to be predicted, i.e. the sequence of labels, as follows:
$$
    (s_L, \dots, s_{L + H}) \rightarrow (e^{\text{output}}_{s_L}, \dots, e^{\text{output}}_{s_{L + H}}).
$$
Similarity scores between the label embeddings and the decoded outputs is computed with a softmax layer using the inner product as a similarity metric:
$$
    \text{score} = \sum_{j = L}^{L + H}
        exp\left(\left<y_j , e^{\text{output}}_{s_j}\right>\right) -
        \log\left(
        \sum_{k = 0 \dots V - 1} exp\left(\left<y_j , e^{\text{output}}_{k}\right>\right)
        \right)
$$
The score corresponds to the log probability of the data prescribed by the parametric RNN model.
Classically, we use a Stochastic Gradient Descent algorithm iterating on an entire data set of sequences to maximize the score w.r.t. to all the paremeters values.
Here the embedding table parameters as well as the RNN and decoding layer are trained.
To accelerate the computation of the softmax during training we employed the standard technique of sampling negatives (see for instance sampled\_softmax\_loss in Tensorflow).

\subsubsection*{Generic EvoRNN architecture for symbol prediction}
The EvoRNN architecture we designed only modifies the recurrent equation~\ref{eq:appendix_rec}.
In all the following, $t$ denotes the position of a symbol w.r.t. to the \emph{beginning} of the input sequence while $t'$ denotes the position w.r.t. to the \emph{end} of the input sequence, i.e. 
$$
    t' = L - 1 - t.
$$
The first step when instantiating an EvoRNN is to create a map from the distance to the end of the input sequence to an RNNCell:
$$
    \text{RNNCellMap}: t, L \rightarrow \text{RNNCellArray}[t' = L - 1 - t] \in {\text{RNNCells}}
$$
which maps a sequential position (counted from the end of the input sequence) to an RNNCell in a set of chosen RNNCells.
$\text{RNNCellArray}$ is an array of pointers to RNNCells.
Typically, a template cell will be chosen such as GRU and the cells in RNNCells will only differ by their number of outputs.

Let us consider a practical example with GRU as the template cell.
We instantiate $4$ different RNNCells with output dimensions $2, 4, 8, 16$: 
$\text{GRU}_2, \text{GRU}_4, \text{GRU}_8, \text{GRU}_{16}$.
To create a ``power-law'' decay in the size of the cell w.r.t. the distance to the end of the sequence, we specify RNNCellMap as follows:
\begin{align*}
    & \text{if } 0 <= L - 1 - t < 2, & RNNCellMap(t, L) & = \text{GRU}_{16} \\
    & \text{if } 2 <= L - 1 - t < 6, & RNNCellMap(t, L) & = \text{GRU}_8 \\
    & \text{if } 6 <= L - 1 - t < 14, & RNNCellMap(t, L) & = \text{GRU}_4 \\
    & \text{if } 14 <= L - 1 - t, & RNNCellMap(t, L) & = \text{GRU}_2. \\
\end{align*}

Once an RNNCellMap has been specified, the RNN recurrent~\ref{eq:appendix_rec} can be modified naively as follows:
\begin{align}
    & h^{i}_0 = \text{init}^i, \\
    & (x^{i + 1}_{t + 1}, h^i_{t + 1}) = 
    \text{RNNCellMap}^i(t, L) (x^i_t, h^i_t).
    \label{eq:appendix_rec_evo_RNN}
\end{align}

Now, as the RNN state dimensions differ between $\text{GRU}_2$ and $\text{GRU}_4$ (being equal to $2$ and $4$ respectively),
the naive EvoRNN equation~\ref{eq:appendix_rec_evo_RNN} is flawed and ill-defined.
To make the RNN state dimensions match, we create a projection map (ProjMap)
$$
    \text{ProjMap}: t \in \left\{0 \dots L - 1\right\} \rightarrow \text{ProjArray}[t] \in {\text{ProjSet}}
$$
that maps a given distance to the end of the input sequence to a projection matrix as follows:
\begin{align*}
    & \text{if } \text{stateDim} \left( \text{RNNCellMap}(t - 1 ,L) \right) \text{ != } \text{stateDim} \left( \text{RNNCellMap}(t, L) \right), \\
    & \text{then } \text{ProjMap}(t, L) = M_t \\
    & \text{else }
    \text{ProjMap}(t, L) = I
\end{align*}
with
$$
M_t \in \mathbb{R}^{
        \text{stateDim} \left( \text{RNNCellMap}(t, L) \right), \text{stateDim} \left( \text{RNNCellMap}(t - 1, L) \right)}
$$
Again, $\text{ProjArray}$ is an array of pointers to real valued matrices of various sizes.

Therefore, the well formed recurrent definition of the EvoRNN writes
\begin{align}
    & h^{i}_0 = \text{init}^i, \\
    & (x^{i + 1}_{t + 1}, h^i_{t + 1}) = 
    \text{RNNCellMap}^i(t, L) (x^i_t, \text{ProjMap}(t, L)(h^i_t)).
    \label{eq:appendix_rec_evo_RNN}
\end{align}

EvoRNN is therefore well defined as a standard RNNCell with the distinction that it needs to be informed, for each input, of its position relative to the end of the sequence.
Such a property is very advantageous as it makes it trivial for EvoRNN to replace any RNN model and work with batched sequences of different lengths.

\subsection*{Implementation of experiments}
We implemented all our experiments in Tensorflow.

\subsubsection*{Implementation of RNN for language prediction experiment}
The baseline implementation corresponds to~\cite{jozefowicz2016exploring} and can be found at

github.com/tensorflow/models/tree/master/research/lm\_1b.

The only architectural modification we made was to the RNN code.
We also slightly changed the training strategy in that only $4$ labels after the end of a $128$ symbol sequence would be used to train and test the model's predictive ability.
The input and output vocabulary sizes are $800$K, $512$ embedding dimensions are used, the number of units in the LSTM is $2048$ and there is only one layer. The decoder is a dense matrix multiplies projecting down to $512$ dimensions (the output embedding dimension).
$8192$ negative samples are employed during training in the softmax layer.
We employ standard Stochastic Gradient descent with a learning rate of $0.2$ and gradient clipping.

\subsubsection*{Implementation of RNN in behavior prediction experiment}
Each input symbol corresponds to $4$ concatenated sub-symbols corresponding respectively to an item embedding, a creator/publisher embedding, a home vs publisher page embedding and finally an interface embedding.
Each input sub-symbol is endowed with its own embedding table.
$128$ dimensions are used to embed items ids, $128$ to embed publishers ids, $32$ to embed the browsed page type and $32$ for the interface type.
There is only one RNN layer comprising a GRU cell with a $256$ dimensional output.
The softmax is trained with $10 000$ negative samples per batch and the sampling distribution we use is a learned unigram distribution (the sampler starts sampling negatives uniformly at random and progressively learns to mimic the historical distribution of unigrams in the data).
The output embedding table comprises $256$ dimensions.
We use a learning rate of $0.1$ with Adagrad and gradient clipping.
The hyper-parameters we use result of careful hyper-parameter tuning of the baseline.
Table~\ref{tab:baseline_youtube_details} gives a summarized view of the critical architectural parameters of the baseline we used on our proprietary data set.

\begin{table}
    \centering
    \begin{tabular}{l|c}
        Parameter & Value \\
        \hline
        Input embedding dimensions & 128 + 128 + 32 + 32 \\
        RNN cell size & 256 \\
        Softmax training negative sampling & Learned unigram \\
        Output embedding dimension & 256 \\
    \end{tabular}
    \caption{\footnotesize{Parameters used in the baseline neural recommender for user behavior. Only the RNN cell is modified and swapped with an EvoRNN. The embeddings given as inputs to the RNN consist of the concatenated embeddings for the video, creator/publisher, interface and page type.}}
    \label{tab:baseline_youtube_details}
\end{table}

\end{document}